\documentclass{article}
\usepackage{ijcai26}

\usepackage{fancyhdr}
\usepackage{times}
\usepackage{soul}
\usepackage{url}
\usepackage[hidelinks]{hyperref}
\usepackage[utf8]{inputenc}
\usepackage[small]{caption}
\usepackage{graphicx}
\usepackage{amsmath}
\usepackage{amsthm}
\usepackage{booktabs}
\usepackage[switch]{lineno}

\usepackage{balance}

\usepackage[utf8]{inputenc} 
\usepackage[T1]{fontenc}    
\usepackage{booktabs}       
\usepackage{amsfonts}       
\usepackage{nicefrac}       
\usepackage{microtype}      
\usepackage{xcolor}         

\usepackage{siunitx}
\DeclareSIUnit{\belmilliwatt}{Bm}
\DeclareSIUnit{\dBm}{\deci\belmilliwatt}

\usepackage{graphicx}
\usepackage{booktabs}

\usepackage{makecell}

\usepackage{tabularx,colortbl}
\usepackage{pifont}

\usepackage{algorithm} 
\usepackage{algpseudocode}

\usepackage{pgfplots}
\usepackage{tikz}

\usepackage{caption}
\usepackage{subcaption}

\usepackage{amsmath}

\usepackage{wrapfig}

\usepackage{multirow}

\usepackage{hhline}

\usepackage{enumitem}

\usepackage[normalem]{ulem}
\usepackage[switch]{lineno}
\usepackage{url}
\usepgfplotslibrary{external}
\usetikzlibrary{shapes}
\usetikzlibrary{positioning}
\usetikzlibrary{arrows,decorations.markings}
\usetikzlibrary{calc,spy,shapes}
\usetikzlibrary{shapes}
\usepgfplotslibrary{fillbetween}

\usetikzlibrary{pgfplots.groupplots}
\usetikzlibrary{patterns}

\newtheorem{remark}{Remark}

\usepackage{xspace}
\usepackage{amsfonts}		
\usepackage[subrefformat=parens]{subcaption}		
\usepackage{siunitx}		
\usepackage{acronym}		

\usepackage{multirow}
\usepackage{nicefrac}       
\usepackage{algorithm}
\usepackage{algpseudocode}
\usepackage{makecell}		
\usepackage{pifont}			

\usepackage[capitalise]{cleveref}		

\graphicspath{{./figures/}}

\usepackage{tikz}
\usepackage{pgfplots}
\usepackage{pgfplotstable}
\usepackage{tikzscale}

\pgfplotsset{compat=1.16}

\pgfplotstableset{search path={.,./figures,./measurements,../measurements}}

\usetikzlibrary{external}
\tikzsetexternalprefix{./plots/cache/}
\usetikzlibrary{positioning}			
\usetikzlibrary{calc}					
\usetikzlibrary{arrows}
\usetikzlibrary{decorations.markings}
\usetikzlibrary{shapes}
\usetikzlibrary{spy}
\usetikzlibrary{pgfplots.groupplots}
\usetikzlibrary{patterns}
\usetikzlibrary{patterns.meta}

\usepgfplotslibrary{fillbetween}

\usepgfplotslibrary{colorbrewer}
\pgfplotsset{
    cycle list/Set1-5,
    cycle multiindex* list={
        mark list*\nextlist
        Set1-5\nextlist,
    },
}

\usepackage{ifthen}

\newboolean{symcheck}
\setboolean{symcheck}{false}

\newboolean{shownotes}
\setboolean{shownotes}{true}

\newboolean{dosavespace}
\setboolean{dosavespace}{false}

\ifthenelse{\boolean{symcheck}}{
	\NewDocumentCommand{\symcheck}{m e{_^}}{%
		{\color{blue!60!black}%
			\IfNoValueTF{#2}{%
				\IfNoValueTF{#3}{#1}{#1^{\color{black}#3}}%
			}{%
				\IfNoValueTF{#3}{#1_{\color{black}#2}}{#1_{{\color{black}#2}}^{\color{black}#3}}%
		}}%
	}
}{
	\newcommand*{\symcheck}[1]{#1}
}

\ifthenelse{\boolean{dosavespace}}{
	
}{

}

\ifthenelse{\boolean{shownotes}}
{
	\usepackage[textsize=footnotesize,linecolor=red,bordercolor=red,backgroundcolor=red!10]{todonotes}	
	\usepackage{setspace}
	\usetikzlibrary{external}
	\setlength{\marginparsep}{2pt}
	\setlength{\marginparwidth}{17.5mm}

	\newcommand{\note}[4]{%
		\ifthenelse{\equal{#1}{footnote}}{{
			\tikzexternaldisable%
			\footnote{#2\todo[caption={},#3,inline]{#4}}%
		}}{{
			\tikzexternaldisable%
			\todo[caption={},#3,#1]{#4}%
		}}{}%
	}

	\newcommand{\draft}[1]{{
		\color{black!30} \vspace{1ex}\hrule\vspace{2pt}
		\color{blue!55!black} #1
		\color{black!30} \vspace{2pt}\hrule\vspace{1ex}
	}}

}
{
	\usepackage[disable]{todonotes}

	\newcommand{\note}[4]{}
	\newcommand{\draft}[1]{}
}

\newcommand{\fakepar}[1]{\paragraph{#1.}}

\newcommand{\capt}[1]{#1}  

\Crefname{section}{Section}{Sections}
\Crefname{figure}{Figure}{Figures}
\Crefname{equation}{Equation}{Equations}

\newcommand*{\eg}{e.g.,\xspace}
\newcommand*{\ie}{i.e.,\xspace}

\DeclareSIUnit{\belmilliwatt}{Bm}
\DeclareSIUnit{\dBm}{\deci\belmilliwatt}

\acrodef{name}[\textsc{CATS}]{\symcheck{Collaborative Inference at the Sensor-level}}
\acused{name}

\acrodef{x-to-all}[x-to-all]{x-to-all}		
\acrodefindefinite{x-to-all}{an}{an}
\acused{x-to-all}

\acrodef{lfa}[LFA]{Low-Frequency Averaging}
\acrodef{mixer}[Mixer]{Mixer} \acused{mixer}
\acrodef{rlnc}[RLNC]{Random Linear Network Coding}
\acrodef{synctrans}[ST]{Syn\-chro\-nous Trans\-mis\-sions}
\acrodef{nn}[NN]{Neural Network}
\acrodef{mlp}[MLP]{Multilayer Perceptron}
\acrodef{butler}[\textsc{Butler}]{} \acused{butler}
\acrodef{wpaxos}[W-Paxos]{Wireless Paxos}
\acrodef{iv}[IV]{InfoVector}
\acrodef{civ}[CIV]{\ac{iv}-based Chaos}

\acrodef{dti}[DTI]{distributed transformer inference}
\acrodef{cnn}[CNN]{convolutional neural network}
\acrodefplural{cnn}[CNNs]{convolutional neural networks}
\acrodef{vit}[ViT]{Vision Transformer}
\acrodefplural{vit}[ViTs]{Vision Transformers}

\acrodef{allgather}[AllGather]{AllGather}
\acused{allgather}
\acrodef{somegather}[SomeGather]{SomeGather}
\acused{somegather}
\acrodef{reducescatter}[ReduceScatter]{ReduceScatter}
\acused{reducescatter}
\acrodef{allreduce}[AllReduce]{AllReduce}
\acused{allreduce}

\acrodef{dropout}[MD]{message-dropout}

\newcommand*{\iNode}{\ensuremath{\symcheck{i}}\xspace}

\newcommand*{\iRound}{\ensuremath{\symcheck{k}}\xspace}

\NewDocumentCommand{\sPrio}{O{\iNode}O{\iRound}}{\ensuremath{\symcheck{g}\IfNoValueF{#1}{_{#1}}\IfNoValueF{#2}{(#2)}}\xspace}

\acrodef{ble}[BLE]{Bluetooth Low Energy}
\acrodef{cfl}[server-based \ac{fl}]{server-based \ac{fl}}		
\acused{cfl}
\acrodef{cps}[CPS]{Cyber-Physical Systems}
\acrodef{dfl}[DFL]{decentralized \ac{fl}}
\acrodef{fl}[FL]{Federated Learning}
\acrodef{i2c}[I${}^2$C]{I${}^2$C}	\acused{i2c}
\acrodef{iid}[i.i.d.]{independent and identically distributed}
\acrodef{iot}[IoT]{Internet of Things}
\acrodef{mcu}[MCU]{microcontroller unit}
\acrodef{ml}[ML]{Machine Learning}	\acrodefindefinite{ml}{an}{a}
\acrodef{ram}[RAM]{Random Access Memory}
\acused{ram}
\acrodef{soc}[SoC]{System on Chip}
\acrodef{uart}[UART]{UART}	\acused{uart}
\acrodef{wifi}[Wi-Fi]{Wi-Fi}	\acused{wifi}

\acrodef{hu2024edge}[Voltage]{Voltage}
\acused{hu2024edge}
\acrodef{liu2025communication}[Astra]{Astra}
\acused{liu2025communication}
\acrodef{bochem2025distributed}[SiracusaDTI]{SiracusaDTI}
\acused{bochem2025distributed}

\newcommand{\speedup}[1]{$#1\times$}

\definecolor{mycolora}{RGB}{31, 119, 180}
\definecolor{mycolorb}{RGB}{174, 199, 232}
\definecolor{mycolorc}{RGB}{255, 127, 14}
\definecolor{mycolord}{RGB}{255, 187, 120}
\definecolor{mycolore}{RGB}{44, 160, 44}
\definecolor{mycolorf}{RGB}{152, 223, 138}
\definecolor{mycolorg}{RGB}{214, 39, 40}
\definecolor{mycolorh}{RGB}{255, 152, 150}
\definecolor{mycolori}{RGB}{148, 103, 189}
\definecolor{mycolorj}{RGB}{197, 176, 213}
\definecolor{mycolork}{RGB}{140, 86, 75}
\definecolor{mycolorl}{RGB}{196, 156, 148}
\definecolor{mycolorm}{RGB}{227, 119, 194}
\definecolor{mycolorn}{RGB}{247, 182, 210}
\definecolor{mycoloro}{RGB}{127, 127, 127}
\definecolor{mycolorp}{RGB}{199, 199, 199}
\definecolor{mycolorq}{RGB}{188, 189, 34}
\definecolor{mycolorr}{RGB}{219, 219, 141}
\definecolor{mycolors}{RGB}{23, 190, 207}
\definecolor{mycolort}{RGB}{158, 218, 229}

\newcommand{\realnmbrs}{\mathbb{R}}

\newcommand{\layernorm}{\mathrm{layernorm}}

\newcommand{\pruningsubmat}[1]{\ensuremath{p}_{#1}}

\title{Going Beyond the Edge: Distributed Inference of Transformer Models on Ultra-Low-Power Wireless Devices}

\author{
Alexander Gräfe$^1$\and
Ding Huo$^2$\and
Vincent de Bakker$^1$\and
Johannes Berger$^1$\and
Marco Zimmerling$^2$\and
Sebastian Trimpe$^1$\\
\affiliations
$^1$RWTH Aachen University\\
$^2$TU Darmstadt\\
\emails
alexander.graefe@dsme.rwth-aachen.de
}

\begin{document}

\fancyhf{}
\fancyhead[C]{\textit{Accepted for Publication at the 35th International Joint Conference on Artificial Intelligence}}
\renewcommand{\headrulewidth}{0pt}

\maketitle
\thispagestyle{fancy}

\begin{abstract}
        
    Transformer models are rapidly becoming a cornerstone of modern \ac{iot} applications, yet their computational and memory demands far exceed the capabilities of a single typical ultra-low-power \ac{iot} device.
    We present \ac{name}, a framework for distributed transformer inference on ultra-low-power wireless devices, enabling multiple devices to collaboratively execute models far larger than what a single device can sustain.
    At its core, \ac{name} is a communication-aware \acl{dti} scheme co-designed across transformer partitioning, wireless communication and training.
    It employs \ac{somegather}, a new pruned communication primitive that selectively broadcasts activation columns to reduce communication bandwidth and \ac{ram} usage without sacrificing model accuracy. 
    Building on \ac{somegather}, we design a partitioning method that exploits this primitive for efficient model parallelism.
    To cope with unreliable wireless communication, \ac{name} 
    employs \acl{dropout} during training, which mimics packet losses and yields models that are robust to 
    message loss during inference.
    In real-world experiments, we show that \ac{name} brings distributed transformer inference to ultra-low-power wireless devices for the first time, with deployments on up to 16 devices that collaboratively execute transformer models up to 14 times larger than what a single device can run.

\end{abstract}

	\section{Introduction}\label{sec:introduction}

Transformer models have become the backbone of intelligent \ac{iot} systems, delivering state-of-the-art performance in natural language processing, computer vision and time-series analysis~\cite{kaur2023survey,zong2025integrating,zhang2025missing}. 
This remarkable capability, however, comes at a substantial computational cost and typically requires powerful hardware, which conflicts with the tight resource budgets of \ac{iot} downstream devices like smart sensors. 

To address these challenges and enable scalable deployment of large transformer models on downstream devices, recent work has identified \ac{dti} as a promising paradigm~\cite{liu2025communication,bochem2025distributed,hu2024edge,zhang2025communication,liu2025efficient}.
In many applications, such as sensor networks, devices naturally operate in groups. 
By pooling resources from multiple devices, \ac{dti} can execute transformers much larger than what a single device can handle,
thereby unlocking advanced transformer models in highly resource-constrained environments. 

Existing approaches, however, target powerful platforms (\eg{} Raspberry Pis or specialized \acp{mcu} with tensor processing units) connected via high-speed wired links. 
In contrast, the vast majority of real-world deployments, such as wireless sensor networks for environmental monitoring, infrastructure or agriculture, are far more constrained~\cite{sanislav2021energy,rahaman2022wireless,sofi2022structural}.
First, they typically rely on small batteries or energy harvesting as a power source and can thus only support ultra-low-power devices, which consist of \acp{mcu} with very limited memory and compute power~\cite{borges2014survey,jamshed2022challenges}, several orders of magnitude lower than that of \eg{} a Raspberry Pi.
Second, devices in such applications communicate via low-power wireless interfaces such as \ac{ble}, forming mesh networks to cover large areas, enable flexible deployment and support mobile nodes~\cite{zimmerling2020,baumann2020wireless,chai2024future}.
While creating unprecedented opportunities for smart \ac{iot} systems, this setting comes with several fundamental challenges for \ac{dti}, which are still unexplored:

\begin{figure}[t]
	\centering
	\includegraphics[width=0.99\linewidth]{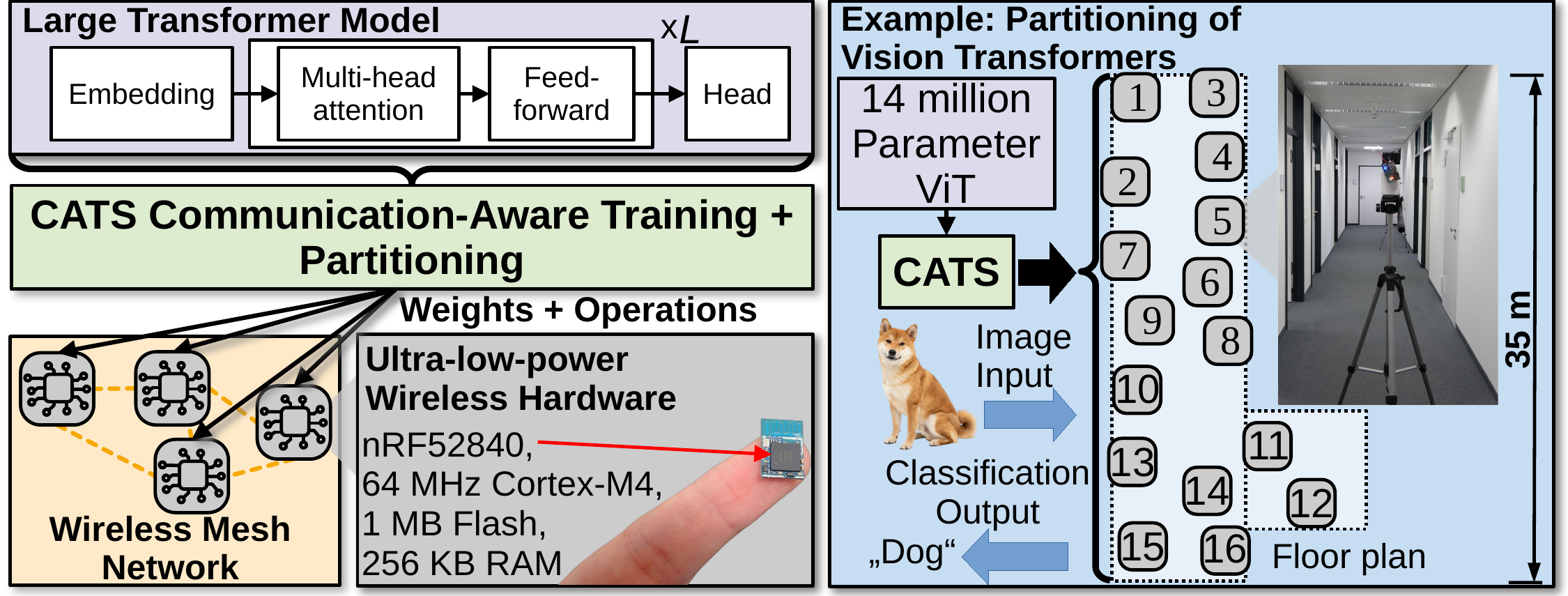}
	\caption{\acf{name}. \capt{\ac{name} contains a communication-aware training and partitioning scheme that enables the execution of large transformer networks to be parallelized across ultra-low-power wireless devices communicating in mesh networks.
	Our resulting hardware implementation consists of 16 nRF52840 devices (\SI{64}{\mega\hertz} Cortex-M4, \qty{1}{\mega\byte} flash, \qty{256}{\kilo\byte} \ac{ram}) spread on a floor of a university building, communicating over a two-hop network on \ac{ble}'s physical layer.
	This implementation is able to execute a 14 million parameter \acf{vit}, fourteen times larger than what a single device could execute.}}
	\label{fig:example}
\end{figure}

\begin{itemize}
	\item[\textbf{C1}] \textbf{Limited Computing Power} \Ac{dti} of transformers on resource-constrained \acp{mcu} is bottlenecked by all key hardware resources: scarce flash memory for storing weights, limited \ac{ram} for intermediate activations and constrained compute throughput for timely inference.
	\item[\textbf{C2}] \textbf{Limited Communication Bandwidth} Low-power wireless \ac{dti} is 
	constrained by low-bandwidth, high-latency inter-device links that turn communication into a dominant bottleneck. 
Mesh networks further exacerbate this challenge as packets are forwarded over multiple hops.
\item[\textbf{C3}] \textbf{Unreliable Communication} 
As wireless communication is inherently unreliable, communicated intermediate activations may be lost, thereby significantly degrading accuracy.
\end{itemize}

\paragraph{} These constraints render existing \ac{dti} approaches~\cite{liu2025communication,bochem2025distributed,hu2024edge,zhang2025communication,liu2025efficient} largely unsuitable in the ultra-low-power wireless setting. 
Many methods overlook critical resource bottlenecks (\textbf{C1}). 
Some reduce \ac{ram} usage but leave flash limits unaddressed~\cite{liu2025communication,hu2024edge}, while others do the opposite~\cite{bochem2025distributed}. 
Communication constraints (\textbf{C2}) are only partially tackled, often by assuming specific network structures, such as grouped device configurations~\cite{bochem2025distributed}, which are unavailable in arbitrary mesh networks and dynamic network topologies. 
Moreover, all approaches assume lossless communication, which may hold in wired settings but fails in wireless networks (\textbf{C3})~\cite{srinivasan2010empirical}. 
As a result, no method for \ac{dti} on ultra-low-power wireless devices exists today, exposing a substantial gap between current \ac{dti} research and the demands of real-world \ac{iot} deployments.

\fakepar{Contributions}
To close this gap, we introduce \ac{name} (\textit{\textbf{C}ol\-laborative Inference \textbf{a}t \textbf{t}he \textbf{S}ensor-level}) (\cref{fig:example}).
Our key idea is to design the transformer partitioning, communication and training jointly around a pruned all-to-all primitive called \ac{somegather}.
By pruning entire activation columns in a communication-aware way, \ac{somegather} turns a dense transformer into one where feature subspaces are processed purely locally, while a carefully chosen subset is shared. 
This simultaneously cuts communication volume and reduces the activations each device must store in \ac{ram} (addresses \textbf{C1} and \textbf{C2}) while maintaining accuracy.
As a result, our partitioning scheme based on \ac{somegather} allows reducing the per-device \ac{ram}, flash and computing load at the same time (\textbf{C1}), unlike existing approaches.
To cope with stochastic message losses (\textbf{C3}), we inject \ac{dropout} during training, a structured dropout exposing the model to realistic loss patterns making it robust to message loss during inference.

As a result, \ac{name} enables efficient collaborative execution of large transformer models across multiple ultra-low-power wireless devices. 
\Cref{fig:example} shows \ac{name} in action on 16 \acp{mcu} communicating via a low-power wireless mesh.
They jointly execute a 14 million parameter \acf{vit}, over $14\times$ larger than fits on a single device.
To the best of our knowledge, this is the first demonstration of \ac{dti} on ultra-low-power wireless hardware, showing that \ac{name} substantially cuts per-device resource requirements and makes large transformers practical on such platforms.
In contrast, current state-of-the-art approaches~\cite{hu2024edge,liu2025communication,bochem2025distributed} deployed on the devices would only execute transformers no larger than what a single device can run.
Experiments further reveal that \ac{somegather} reduces communication volume by up to \SI{90}{\percent} and per-device activation \ac{ram} by \SI{67.5}{\percent} compared to unpruned \ac{dti}, while maintaining accuracy on four time-series prediction benchmarks and \ac{dropout} reduces the relative prediction error increase from none to \SI{10}{\percent} message loss from up to \SI{200}{\percent} to \SI{23.6}{\percent} across datasets.

To summarize, our contributions are:
\begin{enumerate}
	\item A new communication primitive, \ac{somegather}, and a partitioning scheme for transformer models that express all cross-device communication as column-pruned all-to-all exchanges, simultaneously reducing communication, per-device computing load, \ac{ram} and flash footprint.
	\item \Ac{dropout}, a structured dropout mechanism that simulates realistic message-loss patterns across all transformer layers, improving robustness to lossy communication.
	\item The first end-to-end demonstration of \ac{dti} on ultra-low-power wireless devices.
\end{enumerate}

    \section{Background -- Transformer Networks}
\label{sec:background}

This section introduces the notation for transformer models used in this work.
For brevity, we focus on encoder- and decoder-only architectures, which are state-of-the-art in domains such as vision and time-series processing~\cite{dosovitskiy2020image,nie2022time}.
A transformer maps tokens $X_{1}$ (e.g., image or time-series patches~\cite{dosovitskiy2020image,nie2022time}) to new tokens $X_{T}$ via $T$ transformer layers.
Each layer $i$ consists of an attention and a residual block:

\begin{align}
\bar{X}_i &= \text{layernorm}(X_i)\label{eq:transformerbegin} \\
Q_i &= \bar{X}_i W_{i, Q}, \quad
K_i = \bar{X}_i W_{i, K}, \quad
V_i = \bar{X}_i W_{i, V} \\
H_h &= \mathrm{softmax}\!\left(\frac{Q_{i,h}K_{i,h}^T}{\sqrt{F/H}}\right)V_{i,h}, \quad h \in \{1,\dots,H\}\label{eq:selfattention}\\
Y_{i} &= [H_1,H_2,\ldots,H_H]W_{i, O} + X_{i}\\
X_{i+1} &= W_{i, L}f(\ldots W_{i, 2}f(W_{i, 1}\layernorm(Y_{i}))) + Y_{i}.\label{eq:residualblock}
\end{align}

The transformer block first applies row-wise layernorm to its input $X_i$ to obtain $\bar{X}_i$ and then linearly projects it into queries, keys and values $Q_i, K_i, V_i \in \realnmbrs^{N\times F}$ using weight matrices $W_{i,Q}, W_{i,K}, W_{i,V}$ (biases omitted). 
They are split into $H$ parts $Q_{i,h}, K_{i,h}, V_{i,h} \in \realnmbrs^{N\times(F/H)}$, from which attention heads produce $H_h$. 
These are concatenated, projected with weights $W_{i,O}$ and combined with input $X_i$ via a residual connection to obtain the output $Y_i$.
A residual block with $L$ layers, input layernorm, element-wise activation function $f$ and weights $W_{i, \ell}$ then leads the output $X_{i+1}$.

    \section{Related Work}\label{sec:relatedwork}

To our knowledge, no existing methods support \ac{dti} on ultra-low-power wireless devices. 
Current approaches target substantially more powerful edge platforms such as Raspberry Pis~\cite{wei2024communication,liu2025efficient,wen2025easyvit}, Nvidia Jetsons~\cite{xu2023devit}, laptops~\cite{liu2025communication} and virtual machines~\cite{hu2024edge}, and assume wired communication. 
The only study employing \acp{mcu}~\cite{bochem2025distributed} uses Siracusa chips~\cite{prasad2024siracusa} with multiple RISC-V cores, again connected via wired links. 
In contrast, \ac{name} targets cost-effective, common wireless \acp{mcu}, enabling \ac{dti} in an unaddressed regime. 
Given the lack of complete end-to-end solutions, we therefore review prior work that addresses individual aspects relevant to \ac{name}.

\subsection{Partitioning Strategies for Transformers}\label{sec:relatedwork:dti}

\Ac{dti} methods for edge devices differ in their partitioning strategies and which communication primitives they use for them. 

\fakepar{Communication primitives}
Most \ac{dti} methods build on three communication primitives to distribute matrix multiplications: \ac{allgather}, \ac{allreduce} and \ac{reducescatter}~\cite{stahl2021deeperthings,du2024co}.
Using \ac{allgather}, devices initially hold disjoint parts of a matrix and broadcast them so that, afterwards, each device holds the entire matrix. 
Using \ac{allreduce}/\ac{reducescatter}, devices hold partial summands of a global sum and, through collaborative communication, end up with the result of the sum, either the entire matrix (\ac{allreduce}) or only parts (\ac{reducescatter}).
Efficient implementations of \ac{allreduce}/\ac{reducescatter}, however, typically assume specialized, structured communication topologies (e.g., groups~\cite{bochem2025distributed}), and become \ac{ram}- and communication-inefficient in unstructured mesh networks, as we later quantify in \cref{sec:evaluation:memory-scaling}.

\fakepar{Partitioning strategies}
Recent \ac{dti} strategies for edge devices build on these primitives and can be classified into three main categories.
The first class~\cite{liu2025communication,wen2025easyvit,hu2024edge} splits the transformer along the token dimension and communicates during self-attention.
This approach improves computational throughput and reduces \ac{ram} usage but still requires each device to store the full set of weights, leaving the flash footprint unchanged.

The second class partitions the transformer along the feature dimension and distributes different attention heads across devices~\cite{bochem2025distributed}.
This lowers flash memory requirements because each device only stores a subset of the weights. 
However, this approach relies on the communication- and \ac{ram}-heavy \ac{allreduce} primitive.

The third class assigns smaller, independent models to each device and then merges their outputs~\cite{xu2023devit,liu2025efficient}, or decouples transformer components~\cite{wei2024communication}. 
These methods are either restricted to image-classification transformers or ultimately require merging blocks and intermediate results using the previously mentioned strategies, thereby inheriting similar limitations.

\fakepar{Positioning \ac{name}}
\ac{name} builds on the strengths of the second class by dividing the transformer along the feature dimension and distributing attention heads across devices, 
reducing per-device flash memory usage. 
To avoid the communication and \ac{ram} overhead of \ac{allreduce} in mesh networks, \ac{name} relies on \ac{somegather}.
\ac{somegather} is a pruned variant of \ac{allgather} that broadcasts only a selected subset of activation columns, reducing both communication volume and per-device \ac{ram} compared to pure \ac{allgather}.
As \cref{sec:approach:partitioning,sec:approach:pruning} detail, this tailored design simultaneously lowers flash usage, \ac{ram} requirements and per-device computational load, making it well suited for ultra-low-power wireless platforms.

\subsection{Communication-Aware Distributed Inference}\label{sec:relatedwork:mesh}
Current \ac{dti} approaches do not consider constrained and unreliable mesh communication. 
However, methods for other types of \acp{nn} address parts of these challenges.

\fakepar{Communication-aware pruning}
Specialized pruning techniques~\cite{mao2017modnn,abdi2020restructuring,jian2023communication,qin2023disco} reduce communication latency by removing 
transmitted \ac{nn} connections.
While prior work predominantly targets convolutional \acp{nn}, \ac{name} extends this idea to transformers. 
Moreover, we demonstrate that such pruning not only lowers communication overhead but can also reduce \ac{ram} requirements, an aspect that is overlooked in existing communication-aware pruning strategies.

\fakepar{Distributed inference over lossy networks}
Current work on inference under message loss focuses on a two-device, server-edge setting~\cite{itahara2022communication,cheng2024rcif,hou2025loss}, where the edge executes the front part of an \ac{nn} and the server the back part.
These approaches show that simulating lossy communication during training via tailored dropout between the front part and the back part significantly improves robustness at inference time.
\ac{name} generalizes this concept to distributed inference across multiple devices, where the execution of each layer is shared among them. 
We introduce \ac{dropout} that applies dropout simulating lossy communication to \emph{all} layers, going beyond a single split point.

\fakepar{Distributed inference in mesh networks} 
Existing approaches explicitly consider the network graph either in a task-allocation problem~\cite{samikwa2023disnet,jung2023optimization,disabato2021distributed} or during training~\cite{jian2023communication}. 
In realistic wireless settings, the network topology is dynamic, influenced by environmental changes or device mobility (\eg{} wearable devices)~\cite{hao2025dnn}. 
As a result, these approaches require repeated optimization and allocation steps during changes, which incurs substantial overhead. 
\ac{name} follows a different strategy by implementing a communication scheme that abstracts away the details of mesh connectivity while providing properties that are particularly advantageous for \ac{dti} (see \cref{sec:approach:mixer}).

\section{\ac{name} -- Distributed Transformer Inference on Ultra-Low-Power Wireless Devices}\label{sec:approach}

This section presents how \ac{name} enables \ac{dti} on ultra-low-power wireless devices by minimizing per-device \ac{ram}, flash and compute (\textbf{C1}) while addressing limited and unreliable mesh communication (\textbf{C2}, \textbf{C3}). 
We first outline all components and their interactions, then detail each component.

\subsection{Overview}
\label{sec:approach:overview}

We consider $D$ devices in a wireless mesh network with a time-varying topology to accommodate mobile devices like wearables. 
The devices must execute a transformer for data analysis.
We focus on distributing the transformer layer computation (Equations~(\ref{eq:transformerbegin})--(\ref{eq:residualblock})).
Data sources and sinks are application-specific and thus out of scope, but our approach generally supports single as well as distributed sources and sinks.

\begin{figure}[t]
	\centering
	\includegraphics[width=0.99\linewidth]{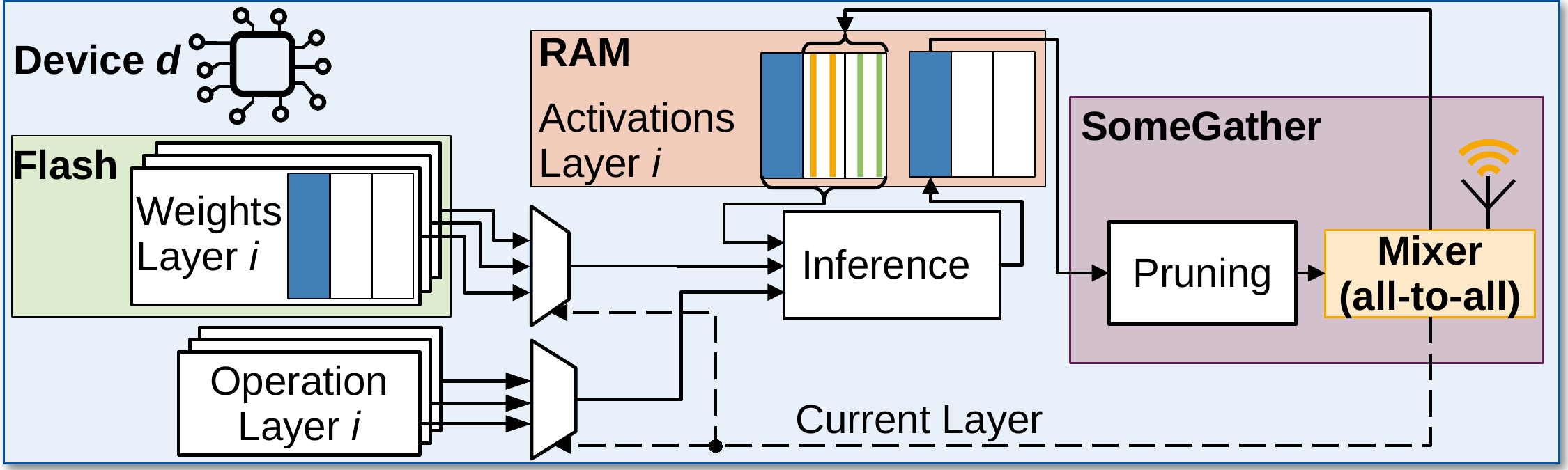}
	\caption{
	Device Operation. \capt{Devices operate in synchronized rounds of communication and computation.
	The \ac{somegather} communication primitive prunes the transmitted data, to reduce bandwidth and \ac{ram} demands.
	Each device stores only a part of the activations and weights, reducing \ac{ram} and flash usage.
	}}
	\label{fig:setup}
\end{figure}

At a high level, \ac{name} organizes inference into tightly synchronized rounds of communication and computation. 
In each round, every device first computes its share of the current layer's operations on its locally stored activations and weights as shown in \cref{fig:setup}. 
The devices then exchange intermediate results using \ac{somegather}, a pruned version of \ac{allgather} that broadcasts only a selected subset of activation columns instead of all of them. 
This allows us to reduce the volume of transmitted data and keep the per-device \ac{ram} for activations small.
On top of this, we design a partitioning scheme that distributes the transformer model across devices such that all inter-device exchanges can be expressed in terms of \ac{somegather} operations.
Finally, \ac{name} uses \ac{dropout}, a mechanism that stochastically removes activations according to Mixer's loss characteristics, making the model robust to packet loss at inference time.
To realize \ac{somegather} efficiently over a wireless mesh, \ac{name} relies on the Mixer protocol~\cite{Glossy,Mixer,zimmerling2020}, which provides all-to-all communication with low latency and high reliability despite dynamic link conditions thus allowing us to abstract mesh communication away during partitioning.

\subsection{\ac{somegather}}\label{sec:approach:pruning}

We illustrate \ac{somegather} on a distributed matrix multiplication $B = AW$ with column-split matrices. 
We consider square matrices $A$ and $W$ with an equal partitioning across devices, \ie{} each device holds $S$ columns of $A$, $W$ and $B$. 
Extensions to non-square matrices or uneven partitionings follow directly by adapting the respective dimensions. 
The next section discusses application of \ac{somegather} to transformers.

In a conventional \ac{allgather}-based implementation, devices first exchange all columns of $A$. 
Each device then multiplies its local copy of $A$ with its local columns of $W$, yielding a subset of the columns of $B = AW$ on every device~\cite{stahl2021deeperthings}.  
In this setting, \ac{ram} scales as $\mathcal{O}(C_\mathrm{A} + C_\mathrm{W}/D)$ and the communication scales as $\mathcal{O}(C_\mathrm{A})$, where $C_\mathrm{A}$ and $C_\mathrm{W}$ denote the sizes of $A$ and $W$.  
This is problematic on ultra-low-power devices, because $C_\mathrm{A}$ can be large.

\Ac{somegather} addresses this bottleneck by pruning communication of activations.  
To support arbitrary sequence lengths $N$, pruning individual rows (\ie{} timestamps) is ineffective. 
We thus prune entire activation columns and broadcast only some columns of $A$ instead of all.
Non-transmitted columns incur no inter-device traffic and are processed purely locally, reducing both the \ac{ram} required for received activations and the overall communication volume.

Mixer's all-to-all behavior constrains how we can prune.
Whenever one device transmits, all others receive the same packet, which precludes pruning individual links between specific pairs of devices, as in prior schemes~\cite{mao2017modnn,abdi2020restructuring,jian2023communication,qin2023disco}.
Pruning must therefore remove all outgoing connections from one device to all other devices.
Formally, each device $d$ maintains a binary matrix $\pruningsubmat{d} \in \{0, 1\}^{S \times 1}$, where an entry of $1$ indicates that the corresponding activation column of $A$ is transmitted.
The pruned multiplication can then be expressed as $B = A (W \odot P)$, with Hadamard product $\odot$ and
\newcommand{\repeatmat}[2]{\ensuremath{(#1)_{#2}}}
\begin{equation}
	P = \begin{bmatrix}
		1_{S\times S} & \repeatmat{\pruningsubmat{1}}{\times S} & \cdots & \repeatmat{\pruningsubmat{1}}{\times S}\\
		\repeatmat{\pruningsubmat{2}}{\times S} & 1_{S\times S} & \cdots & \repeatmat{\pruningsubmat{2}}{\times S}\\
		\cdots & \cdots & \cdots& \cdots\\
		\repeatmat{\pruningsubmat{D}}{\times S} & \repeatmat{\pruningsubmat{D}}{\times S} & \cdots & 1_{S\times S}\\
	\end{bmatrix}.
\end{equation}
where $1_{n \times m}$ is an $n \times m$ matrix of ones and \repeatmat{\pruningsubmat{d}}{\times S} denotes repeating $\pruningsubmat{d}$ $S$ times column-wise. 
The diagonal blocks of $P$ are ones, preserving connections within a device that require no communication. 
The off-diagonal blocks are defined by $\pruningsubmat{d}$, which removes all connections from one device to another. 
The mask $P$ is not used during inference but during training.

Training with pruning is performed stepwise as in~\cite{qin2023disco}. 
At each stage, we prune a fixed fraction of neurons in every layer and then retrain. 
Columns are ranked by the sum of absolute values of all outgoing weights and those with the smallest sums are pruned. 

\subsection{Partitioning Strategy on Top of \ac{somegather}}\label{sec:approach:partitioning}
\begin{figure}[t]
	\centering
	\includegraphics[width=0.99\linewidth]{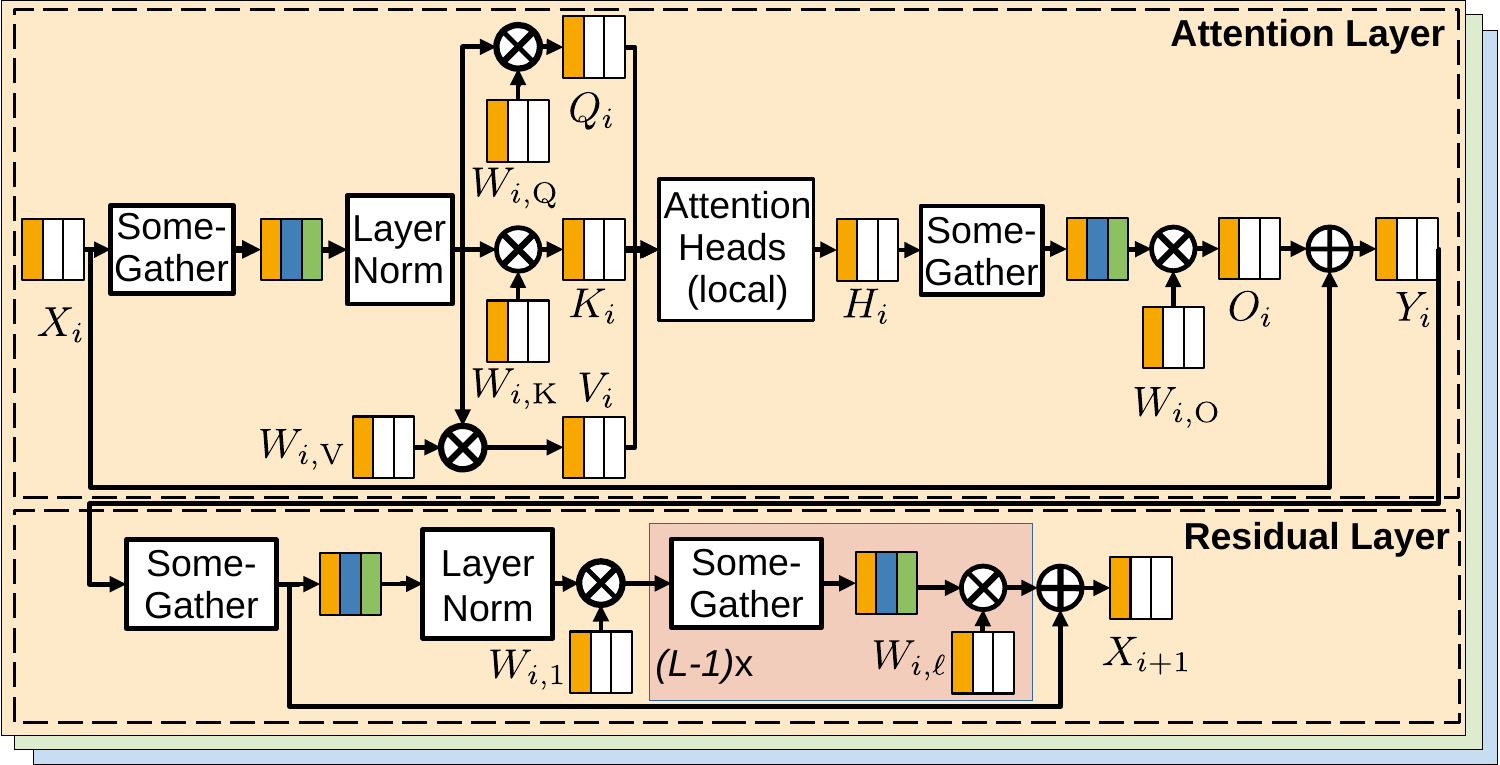}
	\caption{Partitioning Scheme of \ac{name}. \capt{Our partitioning scheme is based on \ac{somegather} and input partitioning of matrix multiplication.
	The multi-head attention mechanism is split along the attention heads, with each device calculating different attention heads.}}
	\label{fig:partitioning}
\end{figure}

Based on this \ac{somegather} primitive, we design a partitioning strategy that relies on it exclusively.
\ac{name} distributes each multi-head attention layer by assigning every device a disjoint subset of the $H$ attention heads and splitting all activations and weights along the feature (column) dimension, as illustrated in \cref{fig:partitioning}.
Each layer starts from an input matrix $X_i$ whose columns are partitioned across devices.
In the first \ac{somegather} operation of the layer, each device receives columns of $X_i$ from all other devices.
The subsequent layernorm requires some adjustments: standard layernorm computes the mean and standard deviation over each complete row, but with pruning no single device necessarily holds all columns of a row. 
We therefore use device-local layernorm, where each device computes the mean and standard deviation from only its activation columns.

After device-local layernorm, devices calculate queries, keys and values locally, using only their assigned columns of $W_{i, Q}, W_{i, K}, W_{i, V}$.
We split the weights such that each device calculates exactly those columns of $Q_i, K_i, V_i$ required by its assigned self-attention heads. 
Consequently, head-specific computations (\cref{eq:selfattention}) are executed fully locally, without further inter-device communication.\footnote{In our implementation, we calculate the heads and their inputs $Q_{i,h}, K_{i,h},V_{i,h}$ sequentially to save \ac{ram}.}
After concatenating their outputs along columns, each device holds distinct columns of $H_i$. 
Those are partially exchanged via a \ac{somegather} operation.
Afterwards, devices multiply it with their assigned columns of $W_{i, O}$ yielding column-wise partitioned $O_i$. 
We partition $W_{i, O}$ so that column indices align with those of the input $X_i$, enabling local addition of the residual connection.  

The subsequent residual block (\cref{eq:residualblock}) follows the same partitioning scheme. 
In each layer $\ell$, devices perform a \ac{somegather} operation followed by multiplication with their designated columns of $W_{i,\ell}$, producing an output also partitioned along columns, serving as input for the next similarly partitioned layer. 
Thus, the entire block's output $X_{i+1}$ is column-partitioned such that it is compatible with the subsequent attention layer's requirements.

\subsection{Message Dropout}\label{sec:approach:dropout}

\ac{dropout} is used to simulate message loss during training.
For Mixer, but also all-to-all protocols in general, we distinguish three modes of message loss:
(a) a device does not receive the message from another device,
(b) a device does not receive messages from any other device,
(c) all other devices do not receive the message from one device.
During training, \ac{name} simulates these three modes by sampling three weight masks.
These masks share a similar structure of the pruning matrix $P$, but are resampled at every training step.
\Ac{dropout} must account for transformer-specific behavior, \ie{} the weight masks for computing queries, keys and values must be identical, and training performs separate layernorm operations for each device.
By sampling such masks at every training step, \ac{name} exposes all layers to realistic patterns of missing messages, which improves robustness under lossy wireless communication.

\subsection{Mixer}
\label{sec:approach:mixer}

\Ac{name} relies on efficient broadcast in wireless mesh networks. 
For our hardware implementation, we therefore employ Mixer~\cite{Mixer}.
Mixer has proven to work under rapid device movement and furthermore offers order-optimal scaling, which is crucial for low communication latency. 
Consequently, \cite{grafe2025rocknet} already demonstrated Mixer's suitability for distributed learning.

Mixer combines synchronous transmissions with random linear network coding to efficiently flood messages through the network~\cite{zimmerling2020,Ho2006}. 
In each \ac{somegather} round, Mixer operates in discrete time slots with microsecond-level synchronization via phase-locked loops, enabling devices to exploit the capture effect~\cite{leentvaar1976capture}.
This effect allows receivers to decode the strongest among simultaneous transmissions, substantially reducing scheduling complexity and communication overhead compared to protocols that avoid concurrent transmissions.
Using network coding, devices transmit linear combinations of their own messages and previously received data, rapidly disseminating information through the network.
By solving the resulting system of equations at the end of each round, all devices can recover all transmitted messages.

    \section{Evaluation}
\label{sec:experiments}

This section presents an in-depth evaluation of \ac{name} and compares it to the state-of-the-art.

\subsection{Evaluation Goals and Methodology}
We evaluate whether \ac{name} addresses challenges \textbf{C1}--\textbf{C3} by structuring our study around four key questions:

\newcommand{\qmemory}[0]{\hyperref[sec:experiments]{\textbf{Q1}}}
\newcommand{\qlatency}[0]{\hyperref[sec:experiments]{\textbf{Q2}}}
\newcommand{\qpruning}[0]{\hyperref[sec:experiments]{\textbf{Q3}}}
\newcommand{\qdropout}[0]{\hyperref[sec:experiments]{\textbf{Q4}}}
\begin{enumerate}
    \item[\qmemory] How does \ac{name} scale regarding memory~(\textbf{C1})?
    \item[\qlatency] How does \ac{name} scale regarding latency (\textbf{C1} and \textbf{C2})?
    \item[\qpruning] Does pruning inside \ac{somegather} preserve model accuracy (\textbf{C1} and \textbf{C2})?
    \item[\qdropout] Can \ac{dropout} make models robust against message loss (\textbf{C3})?
\end{enumerate}

To answer these questions, we combine hardware experiments with simulation studies, providing both realistic validation on actual devices and coverage of a much broader configuration space than feasible on hardware alone.
\footnote{Code available at: \href{https://github.com/Data-Science-in-Mechanical-Engineering/CATS}{github.com/Data-Science-in-Mechanical-Engineering/CATS}.}

\fakepar{Testbed} 
We perform hardware experiments on a testbed consisting of up to 16 devices (nRF52840, 64 MHz Cortex-M4, 1 MB Flash, 256 KB \ac{ram}) placing them along a \SI{35}{\meter} long hallway of a university building leading to a network diameter of at least two hops (see \cref{fig:example}).
The testbed utilizes the original Mixer implementation~\cite{Mixer} using the nRF52840's \ac{ble} physical layer.
For layer computations, we use 8-bit kernels provided by CMSIS-NN~\cite{lai2018cmsis}.

\fakepar{Model and Dataset} Our simulation study uses time-series transformers~\cite{nie2022time} as an exemplary application. 
The models have six transformer layers with a feature dimension of 128 and one hidden layer in the residual blocks.
We train the networks on the ETT-h2~\cite{haoyietal2021informer}, ICD~\cite{von2018self}, London-smart-meters and Traffic~\cite{godahewa2021monash} datasets.
Training uses ADAM~\cite{kingma2014adam} with a batch size of 2048 and a learning rate of $0.001$ with a cosine schedule over 50 epochs.
All runs are repeated ten times with different random seeds.

\fakepar{Metrics} 
We focus on four key system metrics.
\ac{ram} usage and flash footprint for \qmemory{} are obtained analytically.  
Latency for \qlatency{} is measured directly on hardware.
Accuracy for \qpruning{} and \qdropout{}, reported as mean square prediction error, is derived from simulation, since exhaustive hardware evaluation across all models and datasets is infeasible.

\fakepar{Baselines}
As \ac{name} is, to our knowledge, the first method enabling \ac{dti} on ultra-low-power wireless devices, no end-to-end baseline exists.
We therefore use question-specific baselines.
For \qmemory{}, we benchmark our partitioning strategy against \ac{hu2024edge}~\cite{hu2024edge}, \ac{liu2025communication}~\cite{liu2025communication} and \ac{bochem2025distributed}~\cite{bochem2025distributed}, assuming their integration into our Mixer-based framework.
For \qlatency{}, we compare \ac{name}-based distributed execution with centralized execution on a single device.
For \qpruning{}, we compare \ac{somegather} with \ac{allgather} and normal pruning, pruning an \ac{allgather}-distributed transformer to match the bandwidth of a pruned \ac{somegather} model.
For \qdropout{}, we quantify robustness to message loss by comparing models trained with and without \ac{dropout}.

\subsection{\qmemory{} -- Memory Scaling}\label{sec:evaluation:memory-scaling}
We assess \qmemory{} in two experiments.
First, we benchmark against established baselines.
Second, we determine the maximum transformer size that our implementation can execute.

\subsubsection{Baseline Comparison}

\begin{figure}[t]
    \centering

    \pgfplotsset{
        every axis/.append style={
            width=0.3\linewidth,
            height=0.13\linewidth,
            scale only axis,
            xmin=1.0,
            xmax=16,
            xmajorgrids,
            ymajorgrids,
            font=\footnotesize,
            enlarge x limits=0.001,
            legend style={
                at={(2.0, -2.7)},
                anchor=south,
                legend columns=1,
                nodes={anchor=west, scale=0.7, transform shape},
            },
            legend image post style={scale=0.75},
        },
        every axis plot/.append style={line width=1pt},
    }
    \tikzexternaldisable
    \begin{tikzpicture}
        \begin{groupplot}[
            group style={
                group size=2 by 2,
                horizontal sep=1.4cm,
                vertical sep=1.2cm,
            },
        ]
        \nextgroupplot[
            ymin=0,
            ymax=30000,
            xlabel={Number of Devices},
            ylabel={RAM\\ / Bytes},
            ylabel style = {align=center},
        ]
            \addplot[mark=none, color=Dark2-A]
                table [x=num_devices, y=RAM, col sep=comma] {plot_data/resource_CATS.csv};
            \addlegendentry{\ac{name} (Ours)};
            
            \addplot[mark=none, color=Dark2-A, dotted]
                table [x=num_devices, y=RAM, col sep=comma] {plot_data/resource_CATS50.csv};
            \addlegendentry{\ac{name} \SI{50}{\percent} (Ours)};
            
            \addplot[mark=none, color=Dark2-A, dashed]
                table [x=num_devices, y=RAM, col sep=comma] {plot_data/resource_CATS75.csv};
            \addlegendentry{\ac{name} \SI{90}{\percent} (Ours)};
            
            \addplot[mark=none, color=Dark2-B, dashed]
                table [x=num_devices, y=RAM, col sep=comma] {plot_data/resource_Bochem_2025.csv};
            \addlegendentry{\ac{bochem2025distributed}};
            
            \addplot[mark=none, color=Dark2-C,]
                table [x=num_devices, y=RAM, col sep=comma] {plot_data/resource_Hu_2024.csv};
            \addlegendentry{\ac{hu2024edge}};
            
            \addplot[mark=none, color=Dark2-D, dashed]
                table [x=num_devices, y=RAM, col sep=comma] {plot_data/resource_Liu_2025.csv};
            \addlegendentry{\ac{liu2025communication}};

        \nextgroupplot[
            ymin=0,
            ymax=70000,
            xlabel={Number of Devices},
            ylabel={Flash\\ / Bytes},
            ylabel style = {align=center},
        ]
            \addplot[mark=none, color=Dark2-A]
                table [x=num_devices, y=Flash, col sep=comma] {plot_data/resource_CATS.csv};

            \addplot[mark=none, color=Dark2-A, dotted]
                table [x=num_devices, y=Flash, col sep=comma] {plot_data/resource_CATS50.csv};

            \addplot[mark=none, color=Dark2-A, dashed]
                table [x=num_devices, y=Flash, col sep=comma] {plot_data/resource_CATS75.csv};

            \addplot[mark=none, color=Dark2-B, dashed]
                table [x=num_devices, y=Flash, col sep=comma] {plot_data/resource_Bochem_2025.csv};
            
            \addplot[mark=none, color=Dark2-C]
                table [x=num_devices, y=Flash, col sep=comma] {plot_data/resource_Hu_2024.csv};
            
            \addplot[mark=none, color=Dark2-D, dashed]
                table [x=num_devices, y=Flash, col sep=comma] {plot_data/resource_Liu_2025.csv};

        \nextgroupplot[
            ymin=0,
            ymax=70000,
            ylabel={Communication\\ / Bytes},
            ylabel style = {align=center},
            xlabel={Number of Devices},
        ]
            \addplot[mark=none, color=Dark2-A]
                table [x=num_devices, y=Com, col sep=comma] {plot_data/resource_CATS.csv};

            \addplot[mark=none, color=Dark2-A, dotted]
                table [x=num_devices, y=Com, col sep=comma] {plot_data/resource_CATS50.csv};

            \addplot[mark=none, color=Dark2-B, dashed]
                table [x=num_devices, y=Com, col sep=comma] {plot_data/resource_Bochem_2025.csv};
            
            \addplot[mark=none, color=Dark2-C]
                table [x=num_devices, y=Com, col sep=comma] {plot_data/resource_Hu_2024.csv};
            
            \addplot[mark=none, color=Dark2-D, dashed]
                table [x=num_devices, y=Com, col sep=comma] {plot_data/resource_Liu_2025.csv};

            \addplot[mark=none, color=Dark2-A, dashed]
                table [x=num_devices, y=Com, col sep=comma] {plot_data/resource_CATS75.csv};
        \end{groupplot}
    \end{tikzpicture}
    \caption{Resource Usage Comparison Across Different Approaches.  
        \capt{Values following \ac{name} indicate the pruning percentage applied within \ac{somegather}. 
        \ac{name} consistently achieves simultaneous reductions in both \ac{ram} and flash usage, whereas all baseline methods fail to reduce both resources at once.}
        }
    \label{fig:resource-comparison}
\end{figure}
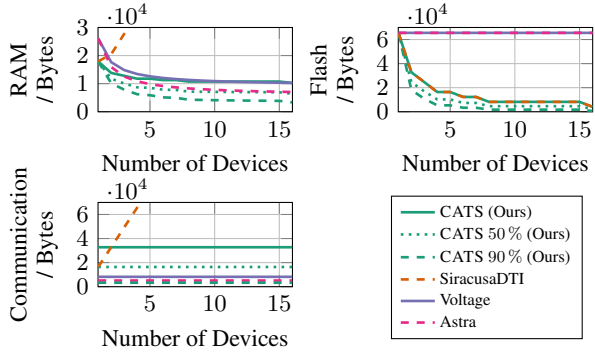

\Cref{fig:resource-comparison} compares \ac{name} against the baselines \ac{liu2025communication}, \ac{hu2024edge} and \ac{bochem2025distributed}.
As the number of devices grows, both \ac{liu2025communication} and \ac{hu2024edge} reduce per-device \ac{ram}. 
\ac{hu2024edge} consistently uses more \ac{ram} than \ac{name}, while \ac{liu2025communication} undercuts unpruned \ac{name} as it relies on specialized low-precision quantization. 
With pruning inside \ac{name}, \ac{ram} drops below \ac{liu2025communication}. 
In contrast, \ac{bochem2025distributed}'s \ac{ram} usage grows linearly with the number of devices due to its \ac{allreduce}-based partitioning, which is poorly supported in mesh networks.

Flash utilization for both \ac{hu2024edge} and \ac{liu2025communication} remains constant, as the weights are not distributed across devices. 
In contrast, \ac{bochem2025distributed} and \ac{name} distribute all weights evenly across devices, resulting in similar Flash usage that systematically decreases with an increasing number of devices.

Without pruning, \ac{name} transmits more data than \ac{hu2024edge} and \ac{liu2025communication}. 
However, by increasing the pruning ratio inside \ac{somegather}, the communication volume can be reduced to comparable levels. 
In contrast, \ac{bochem2025distributed}'s communicated data grows linearly with the number of devices due to its reliance on the \ac{allreduce}-operation.

Overall, no existing baseline reduces both \ac{ram} and flash: \ac{hu2024edge} and \ac{liu2025communication} reduce \ac{ram}, whereas \ac{bochem2025distributed} targets flash. 
In contrast, \ac{name} simultaneously reduces both and its higher communication cost is largely mitigated by pruning in \ac{somegather}, making \ac{name} highly resource-efficient.

\subsubsection{Model Size Scaling}

\begin{figure}[t]
    \centering
    \pgfplotsset{
        every axis/.append style={
                    width=0.35\linewidth,
                    height=0.15\linewidth,
                    scale only axis,
                    ylabel style = {align=center},
                    xlabel style={yshift=2mm},
                    xmajorgrids,
                    ymajorgrids,
                    font=\footnotesize,
                    legend style={
                        at={(0.5, 1.02)},
                        anchor=south,
                        legend columns=2,
                        nodes={anchor=west, scale=0.7, transform shape},
                    },
                    legend image post style={scale=0.75},
                },
                every axis plot/.append style={line width=1.5pt},
    }
    \tikzexternaldisable
    \begin{tikzpicture}
        \begin{groupplot}[
            group style={
                group size=2 by 1,
                horizontal sep=1cm,
                x descriptions at=edge bottom,
            },
            xlabel={$N$},
            ylabel={$F$},    
            xmin=32,
            xmax=1000,
            ymin=0,
            ymax=650,    
            xtick={0, 300, 600, 900},
            title style={font=\fontsize{9pt}{10pt}\selectfont\bfseries, align=center, yshift=0.5cm},
        ]
        \nextgroupplot[
            title={Number of devices}
        ]
                    \addplot[name path=line160, draw=Dark2-A, mark=None, line width=1.5pt]
                      table [x=input_length, y=num_features, col sep=comma] {plot_data/compilation_boundary_16_0.csv};
                    \addlegendentry{16 devices};
                    \addplot[draw=none, name path=axis160, forget plot] coordinates {(32,0) (1500,0)};
                    \addplot[fill=Dark2-A!50!, forget plot] fill between[of=line160 and axis160];

                    \addplot[name path=line80, draw=Dark2-B, mark=None, line width=1.5pt]
                      table [x=input_length, y=num_features, col sep=comma] {plot_data/compilation_boundary_8_0.csv};
                    \addlegendentry{8 devices};
                    \addplot[draw=none, name path=axis80, forget plot] coordinates {(32,0) (1500,0)};
                    \addplot[fill=Dark2-B!50!, forget plot] fill between[of=line80 and axis80];

                    \addplot[name path=line40, draw=Dark2-C, mark=None, line width=1.5pt]
                      table [x=input_length, y=num_features, col sep=comma] {plot_data/compilation_boundary_4_0.csv};
                    \addlegendentry{4 devices};
                    \addplot[draw=none, name path=axis40, forget plot] coordinates {(32,0) (1500,0)};
                    \addplot[fill=Dark2-C!50!, forget plot] fill between[of=line40 and axis40];

                    \addplot[name path=line20, draw=Dark2-D, mark=None, line width=1.5pt]
                      table [x=input_length, y=num_features, col sep=comma] {plot_data/compilation_boundary_2_0.csv};
                    \addlegendentry{2 devices};
                    \addplot[draw=none, name path=axis20, forget plot] coordinates {(32,0) (1500,0)};
                    \addplot[fill=Dark2-D!50!, forget plot] fill between[of=line20 and axis20];
                    
                    \node[anchor=west, font=\scriptsize] (flash) at (axis cs:200,550) {Flash constrained};
                    \draw[->, thick] (flash.west) -- (axis cs:100,520);
                    
                    \node[anchor=west, font=\scriptsize] (ram) at (axis cs:300,400) {\ac{ram} constrained};
                    \draw[->, thick] (ram.west) -- (axis cs:200,370);

        \nextgroupplot[
            title={Pruning},
            ymin=0,
            ymax=1124,   
            ylabel={} 
        ]
            \addplot[name path=line1690, draw=Dark2-A, mark=None, line width=1.5pt]
                table [x=input_length, y=num_features, col sep=comma] {plot_data/compilation_boundary_16_90.csv};
            \addlegendentry{\SI{90}{\percent}};
            \addplot[draw=none, name path=axis1690, forget plot] coordinates {(32,0) (1500,0)};
            \addplot[fill=Dark2-A!50!, forget plot] fill between[of=line1690 and axis1690];

            \addplot[name path=line1660, draw=Dark2-B, mark=None, line width=1.5pt]
                table [x=input_length, y=num_features, col sep=comma] {plot_data/compilation_boundary_16_60.csv};
            \addlegendentry{\SI{60}{\percent}};
            \addplot[draw=none, name path=axis1660, forget plot] coordinates {(32,0) (1500,0)};
            \addplot[fill=Dark2-B!50!, forget plot] fill between[of=line1660 and axis1660];

            \addplot[name path=line1630, draw=Dark2-C, mark=None, line width=1.5pt]
                table [x=input_length, y=num_features, col sep=comma] {plot_data/compilation_boundary_16_30.csv};
            \addlegendentry{\SI{30}{\percent}};
            \addplot[draw=none, name path=axis1630, forget plot] coordinates {(32,0) (1500,0)};
            \addplot[fill=Dark2-C!50!, forget plot] fill between[of=line1630 and axis1630];

            \addplot[name path=line160, draw=Dark2-D, mark=None, line width=1.5pt]
                table [x=input_length, y=num_features, col sep=comma] {plot_data/compilation_boundary_16_0.csv};
            \addlegendentry{\SI{0}{\percent}};
            \addplot[draw=none, name path=axis160, forget plot] coordinates {(32,0) (1500,0)};
            \addplot[fill=Dark2-D!50!, forget plot] fill between[of=line160 and axis160];
        \end{groupplot}
    \end{tikzpicture}
    \caption{Model Size Scaling.
    \capt{Colored regions indicate feasible combinations of feature size $F$ and token count $N$ for each configuration. 
    For small $N$, flash capacity bounds $F$, whereas for larger $N$ the limit is set by \ac{ram}.
    Left plot: Scaling with the number of devices without pruning. Adding devices increases the attainable feature size by decreasing flash usage per device, while \ac{ram} requirements grow only marginally.
    Right plot: Scaling with pruning. Higher pruning ratios (legend) markedly extend the supported token count $N$, as \ac{somegather} substantially reduces the \ac{ram} footprint.}
    }
    \label{fig:memory-scaling}
\end{figure}
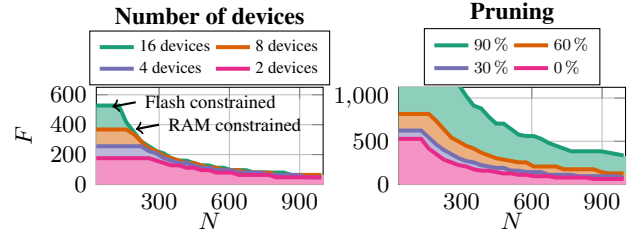

\cref{fig:memory-scaling} quantifies scaling to feature size $F$ and token count $N$. 
For each configuration, we compile the model. If it fails, flash or \ac{ram} limits are exceeded. 
For small $N$, flash capacity bounds $F$, whereas for large $N$, \ac{ram} becomes the dominant bottleneck, since weight flash usage scales quadratically with $F$, while \ac{ram} scales only linearly with $N$.
Increasing the number of devices substantially relaxes flash constraints by distributing the weights onto more devices, while \ac{ram} constraints remain almost unchanged because unpruned \ac{somegather} (\ie{} \ac{allgather}) scales only marginally with the number of devices.
Conversely, increasing pruning in \ac{somegather} reduces flash and \ac{ram} constraints, as devices store fewer activations in \ac{ram} and fewer weights in flash.

In summary, \ac{name} significantly improves model size scalability, with \ac{somegather} as the key mechanism enabling efficient utilization especially of \ac{ram}.

\subsection{\qlatency{} -- Latency Scaling}\label{sec:evaluation:latency-scaling}
\newcommand{\ymaxlargeatt}{4500}
\newcommand{\ymaxlarge}{3500}
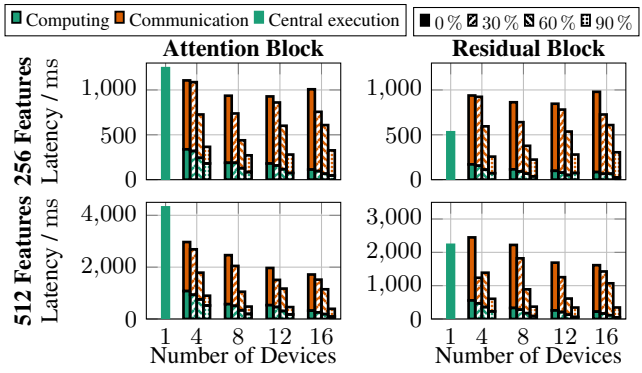
\begin{figure}[t]
  \newcommand{\plotheight}{0.4\linewidth}
  \newcommand{\plotwidth}{0.9\linewidth}
  \centering
  \tikzexternaldisable	
  \tikzsetnextfilename{memory-consumption}
  \pgfplotsset{
  every axis/.append style={
    bar width=2.5pt,
    width=0.3\linewidth,
    height=0.18\linewidth,
    scale only axis,
    xmin=0.0,
    xmax=17.0,
    ymin=0.0,
    ymax=1300,
    enlarge x limits=0.00001,
    xtick={1, 4, 8, 12, 16},
    ylabel style = {align=center},
    xlabel style={yshift=2mm},
    xmajorgrids,
    ymajorgrids,
    font=\footnotesize,
    enlarge x limits=0.05,
    legend style={
        at={(1.7, 1.0)},
        anchor=south,
        legend columns=2,
        nodes={anchor=west, scale=0.7, transform shape},
      },
      legend image post style={scale=0.75},
      },
      every axis plot/.append style={line width=1pt},
      Zero_three/.style={
          pattern={Lines[angle=45, distance=2pt,  line width=0.9pt]}
        },
      Zero_six/.style={
          pattern={Lines[angle=-45, distance=2pt,  line width=0.9pt]}
        },
      Zero_nine/.style={
          pattern={Dots[distance=1.5pt]}
            },
      title style={font=\fontsize{9pt}{10pt}\selectfont\bfseries, align=center, yshift=-0.25cm},
    }
    \tikzexternaldisable
    \begin{tikzpicture}

        \begin{groupplot}[
          ybar stacked,
          /pgf/bar shift=-3.75pt,
            group style={
                group size=2 by 2,
                horizontal sep=1.2cm,
                vertical sep=0.3cm,
                x descriptions at=edge bottom,
            },
        ]
        \nextgroupplot[
            ylabel={\textbf{256 Features}\\Latency / \si{\milli\second}},
            title={Attention Block},
        ]
              \addplot[mark=none, color=black, fill=Dark2-A] coordinates {
            (4 , 338.29)
            (8 , 187.86)
            (12 , 179.94)
            (16, 112.50)
              };
            \label{p1}

          \addplot[mark=none, color=black, fill=Dark2-B] coordinates {
            (4 , 768.099)
            (8 , 747.927)
            (12 , 747.927113702624)
            (16, 895.0631364562119)
              };
            \label{p2}


          \addplot[mark=none, color=black, fill=black] coordinates {
            (4 , 0)
            (8 , 0)
            
          };
          \label{p3}
          \addplot[mark=none, color=black, Zero_three, pattern color=black] coordinates {
            (4 , 0)
            (8 , 0)
            
          };
          \label{p4}

          \addplot[mark=none, color=black, Zero_six, pattern color=black] coordinates {
            (4 , 0)
            (8 , 0)
            
          };
          \label{p5}

          \addplot[mark=none, color=black, Zero_nine, pattern color=black] coordinates {
            (4 , 0)
            (8 , 0)
            
          };
          \label{p6}
       \nextgroupplot[
            title={Residual Block},
        ]
          \addplot[mark=none, color=black, fill=Dark2-A] coordinates {
            (4 , 170.05)
            (8 , 114.97)
            (12, 98.43)
            (16, 84.24)
        };

        \addplot[mark=none, color=black, fill=Dark2-B] coordinates {
             (4 , 768.099)
            (8 , 747.927)
            (12 , 747.927113702624)
            (16, 895.0631364562119)
        };

        \nextgroupplot[
            ylabel={\textbf{512 Features}\\Latency / \si{\milli\second}},
            ymax=\ymaxlargeatt,
            xlabel={Number of Devices},
        ]
              \addplot[mark=none, color=black, fill=Dark2-A] coordinates {
            (4 , 1076.56)
            (8 ,  571.49)
            (12 , 532.89)
            (16, 321.10)
              };
            \label{p1}

          \addplot[mark=none, color=black, fill=Dark2-B] coordinates {
            (4 , 1891.9424460431655)
            (8 , 1891.9424460431655)
            (12 , 1434.6364922206506)
            (16, 1395.1491108071136)
              };
            \label{p2}


          \addplot[mark=none, color=black, fill=black] coordinates {
            (4 , 0)
            (8 , 0)
            
          };
          \label{p3}
          \addplot[mark=none, color=black, Zero_three, pattern color=black] coordinates {
            (4 , 0)
            (8 , 0)
            
          };
          \label{p4}

          \addplot[mark=none, color=black, Zero_six, pattern color=black] coordinates {
            (4 , 0)
            (8 , 0)
            
          };
          \label{p5}

          \addplot[mark=none, color=black, Zero_nine, pattern color=black] coordinates {
            (4 , 0)
            (8 , 0)
            
          };
          \label{p6}
       \nextgroupplot[
            ymax=\ymaxlarge,
            xlabel={Number of Devices},
        ]
          \addplot[mark=none, color=black, fill=Dark2-A] coordinates {
            (4 , 556.31)
            (8 , 330.23)
            (12, 255.88)
            (16, 219.50)
        };

        \addplot[mark=none, color=black, fill=Dark2-B] coordinates {
             (4 , 1891.9424460431655)
            (8 , 1891.9424460431655)
            (12 , 1434.6364922206506)
            (16, 1395.1491108071136)
        };

      \end{groupplot}

      \begin{groupplot}[
        ybar stacked,
        hide axis,
        /pgf/bar shift=-1.25pt,
            group style={
                group size=2 by 2,
                horizontal sep=1.2cm,
                vertical sep=0.3cm,
            },
        ]
        \nextgroupplot[
        ]
              \addplot[mark=none, color=black, Zero_three, pattern color=Dark2-A] coordinates {
            (4 , 319.09)
            (8 , 187.86)
            (12, 157.65)
            (16, 95.41)
              };

              \addplot[mark=none, color=black, Zero_three, pattern color=Dark2-B] coordinates {
            (4 , 768.099)
            (8 , 550)
            (12, 702.6042402826855)
            (16, 660.820253164557)
              };

       \nextgroupplot[
        ]
        \addplot[mark=none, color=black, Zero_three, pattern color=Dark2-A] coordinates {
          (4 , 155.41)
          (8 , 90.86)
          (12, 78.53)
          (16, 65.65)
        };

        \addplot[mark=none, color=black, Zero_three, pattern color=Dark2-B] coordinates {
          (4 , 768.099)
          (8 , 550)
          (12, 702.6042402826855)
          (16, 660.820253164557)
          
        };

        \nextgroupplot[
          ymax=\ymaxlargeatt,
        ]
              \addplot[mark=none, color=black, Zero_three, pattern color=Dark2-A] coordinates {
            (4 , 933.89)
            (8 , 507.37)
            (12, 462.59)
            (16, 253.93)
              };

              \addplot[mark=none, color=black, Zero_three, pattern color=Dark2-B] coordinates {
            (4 , 1754.3702359346642)
            (8 , 1539.8404452690168)
            (12, 1046.0948766603415)
            (16, 1263.747680890538)
              };

       \nextgroupplot[
        ymax=\ymaxlarge,
        ]
        \addplot[mark=none, color=black, Zero_three, pattern color=Dark2-A] coordinates {
          (4 , 469.48)
          (8 , 280.42)
          (12, 208.34)
          (16, 162.78)
        };

        \addplot[mark=none, color=black, Zero_three, pattern color=Dark2-B] coordinates {
          (4 , 768.099)
          (8 , 1539.8404452690168)
          (12, 1046.0948766603415)
          (16, 1263.747680890538)
          
        };

      \end{groupplot}

      \begin{groupplot}[
        hide axis,
            ybar stacked,
        /pgf/bar shift=1.25pt,
            group style={
                group size=2 by 2,
                horizontal sep=1.2cm,
                vertical sep=0.3cm,
            },
        ]
        \nextgroupplot[
        ]
              \addplot[mark=none, color=black, Zero_six, pattern color=Dark2-A] coordinates {
            (4 , 243.18)
            (8 , 128.18)
            (12, 116.81)
            (16, 66.22)
              };

              \addplot[mark=none, color=black, Zero_six, pattern color=Dark2-B] coordinates {
            (4 , 483.585)
            (8 , 310.701)
            (12, 484.20)
            (16, 542.2962536023055)
              };

       \nextgroupplot[
        ]
        \addplot[mark=none, color=black, Zero_six, pattern color=Dark2-A] coordinates {
          (4 , 110.92)
          (8 , 66.90)
          (12, 52.22)
          (16, 66.22)
        };

        \addplot[mark=none, color=black, Zero_six, pattern color=Dark2-B] coordinates {
            (4 , 483.585)
            (8 , 310.701)
            (12, 484.20)
            (16, 542.2962536023055)
        };

        \nextgroupplot[
          ymax=\ymaxlargeatt,
        ]
              \addplot[mark=none, color=black, Zero_six, pattern color=Dark2-A] coordinates {
            (4 , 757.04)
            (8 , 336.15)
            (12, 307.41)
            (16, 187.92)
              };

              \addplot[mark=none, color=black, Zero_six, pattern color=Dark2-B] coordinates {
            (4 , 1030.5164835164835)
            (8 , 715.3565310492505)
            (12, 865.3453724604966)
            (16, 958.37)
              };

       \nextgroupplot[
        ymax=\ymaxlarge,
        ]
        \addplot[mark=none, color=black, Zero_six, pattern color=Dark2-A] coordinates {
          (4 , 358.34)
          (8 , 174.26)
          (12, 129.25)
          (16, 112.37)
        };

        \addplot[mark=none, color=black, Zero_six, pattern color=Dark2-B] coordinates {
            (4 , 1030.5164835164835)
            (8 , 715.3565310492505)
            (12, 484.20)
            (16, 958.37)
        };

      \end{groupplot}

      \begin{groupplot}[
            ybar stacked,
        hide axis,
        /pgf/bar shift=3.75pt,
            group style={
                group size=2 by 2,
                horizontal sep=1.2cm,
                vertical sep=0.3cm,
            },
        ]
        \nextgroupplot[
        ]

              \addplot[mark=none, color=black, Zero_nine, pattern color=Dark2-A] coordinates {
            (4 , 180.98)
            (8 , 84.83)
            (12, 74.28)
            (16, 46.13)
              };

              \addplot[mark=none, color=black, Zero_nine, pattern color=Dark2-B] coordinates {
            (4 , 184.97)
            (8 , 186.662)
            (12, 205.82270916334662)
            (16, 280.66733067729086)
              };

       \nextgroupplot[
        ]

        \addplot[mark=none, color=black, Zero_nine, pattern color=Dark2-A] coordinates {
          (4 , 71.94)
          (8 , 37.07)
          (12, 74.28)
          (16, 24.65)
        };

        \addplot[mark=none, color=black, Zero_nine, pattern color=Dark2-B] coordinates {
            (4 , 184.97)
            (8 , 186.662)
            (12, 205.82270916334662)
            (16, 280.66733067729086)
        };

        \nextgroupplot[
          ymax=\ymaxlargeatt,
        ]

              \addplot[mark=none, color=black, Zero_nine, pattern color=Dark2-A] coordinates {
            (4 , 521.03)
            (8 , 195.66)
            (12, 175.99)
            (16, 93.76)
              };

              \addplot[mark=none, color=black, Zero_nine, pattern color=Dark2-B] coordinates {
            (4 , 377.5345454545455)
            (8 , 283.1454545454545)
            (12, 283.1864111498258)
            (16, 299)
              };

       \nextgroupplot[
        ymax=\ymaxlarge,
        ]

        \addplot[mark=none, color=black, Zero_nine, pattern color=Dark2-A] coordinates {
          (4 , 228.43)
          (8 , 86.81)
          (12, 60.40)
          (16, 46.20)
        };

        \addplot[mark=none, color=black, Zero_nine, pattern color=Dark2-B] coordinates {
            (4 , 377.5345454545455)
            (8 , 283.1454545454545)
            (12, 283.1864111498258)
            (16, 299)
        };

      \end{groupplot}

        \begin{groupplot}[
          hide axis,
          ybar stacked,
          group style={
              group size=2 by 2,
              horizontal sep=1.2cm,
              vertical sep=0.3cm,
          },
          ]
          \nextgroupplot

            \addplot+[mark=none, color=Dark2-A] coordinates {
          (1, 1242.69)
            };
            \label{p7}

         \nextgroupplot

          \addplot+[mark=none, color=Dark2-A] coordinates {
            (1,528.24)
          };

          \nextgroupplot[
            ymax=\ymaxlargeatt,
          ]

            \addplot+[mark=none, color=Dark2-A] coordinates {
          (1, 4306.24)
            };
            \label{p7}

         \nextgroupplot[
          ymax=\ymaxlarge,
         ]

          \addplot+[mark=none, color=Dark2-A] coordinates {
            (1,2225.24)
          };
        \end{groupplot}


      \node [draw,fill=white, scale=0.7,anchor=south] at (rel axis cs: 0.35, 1.23) {\shortstack[l]{
      \ref{p1} Computing
      \ref{p2} Communication
      \ref{p7} Central execution
      }};

      \node [draw,fill=white, scale=0.7, ,anchor=south] at (rel axis cs: 2.0, 1.23) {\shortstack[l]{
      \ref{p3} \SI{0}{\percent}
      \ref{p4} \SI{30}{\percent}
      \ref{p5} \SI{60}{\percent}
      \ref{p6} \SI{90}{\percent}
      }};
    \end{tikzpicture}

    \caption{Latency of \ac{name} Versus Number of Devices. 
    \capt{We measure latency for attention and residual blocks for 256 and 512 features. 
    Different bar patterns show different \ac{somegather} pruning ratios. 
    Central execution latency for 512 features is estimated by multiplying the computing latency at four devices by four as the layers do not fit on a single device.
    Computing latency decreases with number of devices while communication latency stays approximately constant.
    Pruning drastically speeds up execution. 
    }}
  \label{fig:experiments:latency}
\end{figure}

Using the hardware implementation of \ac{name}, we evaluate the latency of attention and residual blocks (\cref{fig:experiments:latency}). 
We first focus on non-pruned inference (solid bars in \cref{fig:experiments:latency}). 
Computing latency decreases with more devices, whereas communication latency remains nearly constant (with a slight increase for 16 devices and 256 features) and dominates runtime, ranging from \SI{69.42}{\percent} (attention block with four devices) to \SI{91.4}{\percent} (residual block with 16 devices). 
Despite this, \ac{name} achieves a speedup of \speedup{1.27} for the attention block at 256 features and \speedup{2.51} at 512 features compared to central execution. 
For the residual block, \ac{name} shows a slowdown at 256 features (\speedup{0.54}), but reaches a speedup of \speedup{1.38} at 512 features.

Pruning via \ac{somegather} markedly enhances scalability.
With a pruning ratio of \SI{90}{\percent}, communication latency is reduced by up to \SI{80}{\percent}. 
For 16 devices, increasing pruning from \SI{0}{\percent} to \SI{90}{\percent} accelerates the attention block by \speedup{3.08} at 256 features and \speedup{4.37} at 512 features and the residual block by \speedup{3.21} and \speedup{4.68}, respectively. 
Overall, pruned inference latency scales from central execution to 16 devices by factors of \speedup{3.8} (256 features) and \speedup{10.96} (512 features) for the attention block and \speedup{1.73} and \speedup{6.45} for the residual block.

\ac{name} thus parallelizes computation efficiently.
Together with \ac{somegather}'s pruning, \ac{name} delivers scalable acceleration despite communication latency.

\begin{remark}
    The reported latencies depend strongly on the underlying hardware: faster radios and slower processors increase speedups and vice versa. Nevertheless, the fundamental behavior of \ac{name} and \ac{somegather} remain unchanged.
\end{remark}

\subsection{\qpruning{} -- Accuracy of \ac{somegather}}\label{sec:evaluation:accuracy-pruning}
\begin{figure}[t]
    \centering
    \pgfplotsset{
        every axis/.append style={
                    width=0.32\linewidth,
                    height=0.14\linewidth,
                    scale only axis,
                    xmin=0.0,
                    xmax=100,
                    enlarge x limits=0.00001,
                    xtick={0.0, 25, 50, 75, 100},
                    ylabel style = {align=center},
                    xlabel style={yshift=2mm},
                    xlabel={Communication savings / \si{\percent}},
                    xmajorgrids,
                    ymajorgrids,
                    font=\footnotesize,
                    legend style={
                        at={(1.3, 1.05)},
                        anchor=south,
                        legend columns=1,
                        nodes={anchor=west, scale=0.7, transform shape},
                    },
                    legend image post style={scale=0.75},
                },
                every axis plot/.append style={line width=1pt},
                title style={font=\fontsize{9pt}{10pt}\selectfont\bfseries, align=center, yshift=-0.25cm},
    }
    \tikzexternaldisable
    \begin{tikzpicture}
        \begin{groupplot}[
            group style={
                group size=2 by 2,
                horizontal sep=1.5cm,
                vertical sep=0.35cm,
                x descriptions at=edge bottom,
            },
        ]
        \nextgroupplot[
            ylabel={Loss},
            title={ETT},
        ]
            \addplot[name path=max, draw=none, forget plot]
                table [x=communication_savings, y=max_loss, col sep=comma] {plot_data/ETT_inter_device_pruning_vs_communication.csv};
            \addplot[name path=min, draw=none, forget plot]
                table [x=communication_savings, y=min_loss, col sep=comma] {plot_data/ETT_inter_device_pruning_vs_communication.csv};
            \addplot[fill=Dark2-A, fill opacity=0.3, draw=none, forget plot] fill between[of=max and min];
            
            \addplot[mark=o, color=Dark2-A, mark size=2pt, line width=1.5pt]
                table [x=communication_savings, y=mean_loss, col sep=comma] {plot_data/ETT_inter_device_pruning_vs_communication.csv};
            \label{ref:somegather}

            \addplot[name path=max, draw=none, forget plot]
                table [x=communication_savings, y=max_loss, col sep=comma] {plot_data/ETT_normal_pruning_vs_communication.csv};
            \addplot[name path=min, draw=none, forget plot]
                table [x=communication_savings, y=min_loss, col sep=comma] {plot_data/ETT_normal_pruning_vs_communication.csv};
            \addplot[fill=Dark2-B, fill opacity=0.3, draw=none, forget plot] fill between[of=max and min];
            
            \addplot[mark=square, color=Dark2-B, mark size=2pt, line width=1.5pt]
                table [x=communication_savings, y=mean_loss, col sep=comma] {plot_data/ETT_normal_pruning_vs_communication.csv};
            \label{ref:normal-pruning}

        \nextgroupplot[
            title={ICD},
        ]
            \addplot[name path=max, draw=none, forget plot]
                table [x=communication_savings, y=max_loss, col sep=comma] {plot_data/ICD_inter_device_pruning_vs_communication.csv};
            \addplot[name path=min, draw=none, forget plot]
                table [x=communication_savings, y=min_loss, col sep=comma] {plot_data/ICD_inter_device_pruning_vs_communication.csv};
            \addplot[fill=Dark2-A, fill opacity=0.3, draw=none] fill between[of=max and min];

            \addplot[mark=o, color=Dark2-A, mark size=2pt, line width=1.5pt]
                table [x=communication_savings, y=mean_loss, col sep=comma] {plot_data/ICD_inter_device_pruning_vs_communication.csv};

            \addplot[name path=max, draw=none, forget plot]
                table [x=communication_savings, y=max_loss, col sep=comma] {plot_data/ICD_normal_pruning_vs_communication.csv};
            \addplot[name path=min, draw=none, forget plot]
                table [x=communication_savings, y=min_loss, col sep=comma] {plot_data/ICD_normal_pruning_vs_communication.csv};
            \addplot[fill=Dark2-B, fill opacity=0.3, draw=none, forget plot] fill between[of=max and min];
            
            \addplot[mark=square, color=Dark2-B, mark size=2pt, line width=1.5pt]
                table [x=communication_savings, y=mean_loss, col sep=comma] {plot_data/ICD_normal_pruning_vs_communication.csv};

        \nextgroupplot[
            ylabel={Loss},
            title={Traffic},
            title style={yshift=-0.04cm},
        ]
            \addplot[name path=max, draw=none, forget plot]
                table [x=communication_savings, y=max_loss, col sep=comma] {plot_data/traffic_hourly_dataset_inter_device_pruning_vs_communication.csv};
            \addplot[name path=min, draw=none, forget plot]
                table [x=communication_savings, y=min_loss, col sep=comma] {plot_data/traffic_hourly_dataset_inter_device_pruning_vs_communication.csv};
            \addplot[fill=Dark2-A, fill opacity=0.3, draw=none] fill between[of=max and min];

            \addplot[mark=o, color=Dark2-A, mark size=2pt, line width=1.5pt]
                table [x=communication_savings, y=mean_loss, col sep=comma] {plot_data/traffic_hourly_dataset_inter_device_pruning_vs_communication.csv};

            \addplot[name path=max, draw=none, forget plot]
                table [x=communication_savings, y=max_loss, col sep=comma] {plot_data/traffic_hourly_dataset_normal_pruning_vs_communication.csv};
            \addplot[name path=min, draw=none, forget plot]
                table [x=communication_savings, y=min_loss, col sep=comma] {plot_data/traffic_hourly_dataset_normal_pruning_vs_communication.csv};
            \addplot[fill=Dark2-B, fill opacity=0.3, draw=none, forget plot] fill between[of=max and min];
            
            \addplot[mark=square, color=Dark2-B, mark size=2pt, line width=1.5pt]
                table [x=communication_savings, y=mean_loss, col sep=comma] {plot_data/traffic_hourly_dataset_normal_pruning_vs_communication.csv};

        \nextgroupplot[
            title={London-smart-meters},
        ]
            \addplot[name path=max, draw=none, forget plot]
                table [x=communication_savings, y=max_loss, col sep=comma] {plot_data/london_smart_meters_dataset_inter_device_pruning_vs_communication.csv};
            \addplot[name path=min, draw=none, forget plot]
                table [x=communication_savings, y=min_loss, col sep=comma] {plot_data/london_smart_meters_dataset_inter_device_pruning_vs_communication.csv};
            \addplot[fill=Dark2-A, fill opacity=0.3, draw=none] fill between[of=max and min];

            \addplot[mark=o, color=Dark2-A, mark size=2pt, line width=1.5pt]
                table [x=communication_savings, y=mean_loss, col sep=comma] {plot_data/london_smart_meters_dataset_inter_device_pruning_vs_communication.csv};

            \addplot[name path=max, draw=none, forget plot]
                table [x=communication_savings, y=max_loss, col sep=comma] {plot_data/london_smart_meters_dataset_normal_pruning_vs_communication.csv};
            \addplot[name path=min, draw=none, forget plot]
                table [x=communication_savings, y=min_loss, col sep=comma] {plot_data/london_smart_meters_dataset_normal_pruning_vs_communication.csv};
            \addplot[fill=Dark2-B, fill opacity=0.3, draw=none, forget plot] fill between[of=max and min];
            
            \addplot[mark=square, color=Dark2-B, mark size=2pt, line width=1.5pt]
                table [x=communication_savings, y=mean_loss, col sep=comma] {plot_data/london_smart_meters_dataset_normal_pruning_vs_communication.csv};
        \end{groupplot}
    \end{tikzpicture}
    \caption{Comparison of Normal Pruning (\ref{ref:normal-pruning}) and \ac{somegather} (\ref{ref:somegather}) on Accuracy Versus Communication Trade-off. \capt{\ac{somegather}'s accuracy remains approximately constant with decreasing communication, 
    whereas normal pruning's accuracy degrades significantly.}}
    \label{fig:pruning-comparison}
\end{figure}
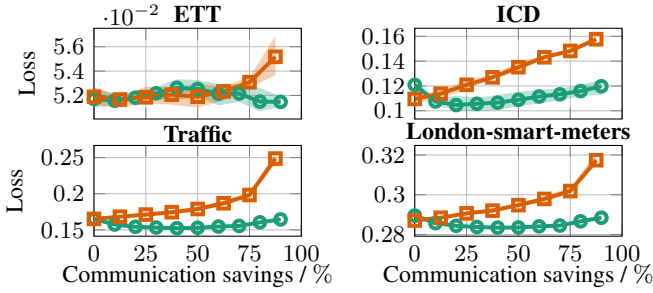

\Cref{fig:pruning-comparison} compares \ac{somegather} with normal pruning in terms of prediction error versus communication savings. 
For normal pruning, prediction error rises steadily as communication is reduced, whereas \ac{somegather} maintains an almost constant error across the entire range. 
Thus, the gains reported in \cref{sec:evaluation:memory-scaling,sec:evaluation:latency-scaling} are achieved without sacrificing accuracy.

\subsection{\qdropout{} -- Robustness Against Message Loss}\label{sec:evaluation:message-loss}
\begin{figure}[t]
    \centering
    \pgfplotsset{
        every axis/.append style={
                    width=0.32\linewidth,
                    height=0.14\linewidth,
                    scale only axis,
                    xmin=0.0,
                    xmax=10,
                    enlarge x limits=0.00001,
                    xtick={0.0, 2.5, 5.0, 7.5, 10},
                    ylabel style = {align=center},
                    xlabel style={yshift=2mm},
                    xlabel={Message loss / \si{\percent}},
                    xmajorgrids,
                    ymajorgrids,
                    font=\footnotesize,
                    legend style={
                        at={(2.3, 0.5)},
                        anchor=west,
                        legend columns=1,
                        nodes={anchor=west, scale=0.7, transform shape},
                    },
                    legend image post style={scale=0.75},
                },
                every axis plot/.append style={line width=1pt},
                title style={font=\fontsize{9pt}{10pt}\selectfont\bfseries, align=center, yshift=-0.25cm},
    }
    \tikzexternaldisable
    \begin{tikzpicture}
        \begin{groupplot}[
            group style={
                group size=2 by 2,
                horizontal sep=1.5cm, 
                vertical sep=0.35cm,
                x descriptions at=edge bottom,
            },
        ]
        \nextgroupplot[
            ylabel={Loss},
            title={ETT},
        ]
            \addplot[name path=max, draw=none, forget plot]
                table [x=message_loss_prob, y=max_loss, col sep=comma] {plot_data/ETT_dropout_0.csv};
            \addplot[name path=min, draw=none, forget plot]
                table [x=message_loss_prob, y=min_loss, col sep=comma] {plot_data/ETT_dropout_0.csv};
            \addplot[fill=Dark2-A, fill opacity=0.3, draw=none, forget plot] fill between[of=max and min];
            \addplot[mark=None, color=Dark2-A, mark size=2pt, line width=1.5pt]
                table [x=message_loss_prob, y=mean_loss, col sep=comma] {plot_data/ETT_dropout_0.csv};
            \label{ref:zero}

            \addplot[name path=max, draw=none, forget plot]
                table [x=message_loss_prob, y=max_loss, col sep=comma] {plot_data/ETT_dropout_1.csv};
            \addplot[name path=min, draw=none, forget plot]
                table [x=message_loss_prob, y=min_loss, col sep=comma] {plot_data/ETT_dropout_1.csv};
            \addplot[fill=Dark2-B, fill opacity=0.3, draw=none, forget plot] fill between[of=max and min];
            \addplot[mark=None, color=Dark2-B, mark size=2pt, line width=1.5pt]
                table [x=message_loss_prob, y=mean_loss, col sep=comma] {plot_data/ETT_dropout_1.csv};
            \label{ref:one}

            \addplot[name path=max, draw=none, forget plot]
                table [x=message_loss_prob, y=max_loss, col sep=comma] {plot_data/ETT_dropout_5.csv};
            \addplot[name path=min, draw=none, forget plot]
                table [x=message_loss_prob, y=min_loss, col sep=comma] {plot_data/ETT_dropout_5.csv};
            \addplot[fill=Dark2-C, fill opacity=0.3, draw=none, forget plot] fill between[of=max and min];
            \addplot[mark=None, color=Dark2-C, mark size=2pt, line width=1.5pt]
                table [x=message_loss_prob, y=mean_loss, col sep=comma] {plot_data/ETT_dropout_5.csv};
            \label{ref:five}

            \addplot[name path=max, draw=none, forget plot]
                table [x=message_loss_prob, y=max_loss, col sep=comma] {plot_data/ETT_dropout_10.csv};
            \addplot[name path=min, draw=none, forget plot]
                table [x=message_loss_prob, y=min_loss, col sep=comma] {plot_data/ETT_dropout_10.csv};
            \addplot[fill=Dark2-D, fill opacity=0.3, draw=none, forget plot] fill between[of=max and min];
            \addplot[mark=None, color=Dark2-D, mark size=2pt, line width=1.5pt]
                table [x=message_loss_prob, y=mean_loss, col sep=comma] {plot_data/ETT_dropout_10.csv};
            \label{ref:ten}

        \nextgroupplot[
            title={ICD},
        ]
            \addplot[name path=max, draw=none, forget plot]
                table [x=message_loss_prob, y=max_loss, col sep=comma] {plot_data/ICD_dropout_0.csv};
            \addplot[name path=min, draw=none, forget plot]
                table [x=message_loss_prob, y=min_loss, col sep=comma] {plot_data/ICD_dropout_0.csv};
            \addplot[fill=Dark2-A, fill opacity=0.3, draw=none, forget plot] fill between[of=max and min];
            \addplot[mark=None, color=Dark2-A, mark size=2pt, line width=1.5pt]
                table [x=message_loss_prob, y=mean_loss, col sep=comma] {plot_data/ICD_dropout_0.csv};

            \addplot[name path=max, draw=none, forget plot]
                table [x=message_loss_prob, y=max_loss, col sep=comma] {plot_data/ICD_dropout_1.csv};
            \addplot[name path=min, draw=none, forget plot]
                table [x=message_loss_prob, y=min_loss, col sep=comma] {plot_data/ICD_dropout_1.csv};
            \addplot[fill=Dark2-B, fill opacity=0.3, draw=none, forget plot] fill between[of=max and min];
            \addplot[mark=None, color=Dark2-B, mark size=2pt, line width=1.5pt]
                table [x=message_loss_prob, y=mean_loss, col sep=comma] {plot_data/ICD_dropout_1.csv};

            \addplot[name path=max, draw=none, forget plot]
                table [x=message_loss_prob, y=max_loss, col sep=comma] {plot_data/ICD_dropout_5.csv};
            \addplot[name path=min, draw=none, forget plot]
                table [x=message_loss_prob, y=min_loss, col sep=comma] {plot_data/ICD_dropout_5.csv};
            \addplot[fill=Dark2-C, fill opacity=0.3, draw=none, forget plot] fill between[of=max and min];
            \addplot[mark=None, color=Dark2-C, mark size=2pt, line width=1.5pt]
                table [x=message_loss_prob, y=mean_loss, col sep=comma] {plot_data/ICD_dropout_5.csv};

            \addplot[name path=max, draw=none, forget plot]
                table [x=message_loss_prob, y=max_loss, col sep=comma] {plot_data/ICD_dropout_10.csv};
            \addplot[name path=min, draw=none, forget plot]
                table [x=message_loss_prob, y=min_loss, col sep=comma] {plot_data/ICD_dropout_10.csv};
            \addplot[fill=Dark2-D, fill opacity=0.3, draw=none, forget plot] fill between[of=max and min];
            \addplot[mark=None, color=Dark2-D, mark size=2pt, line width=1.5pt]
                table [x=message_loss_prob, y=mean_loss, col sep=comma] {plot_data/ICD_dropout_10.csv};

        \nextgroupplot[
            ylabel={Loss},
            title={Traffic},
            title style={yshift=-0.04cm},
        ]
            \addplot[name path=max, draw=none, forget plot]
                table [x=message_loss_prob, y=max_loss, col sep=comma] {plot_data/traffic_hourly_dataset_dropout_0.csv};
            \addplot[name path=min, draw=none, forget plot]
                table [x=message_loss_prob, y=min_loss, col sep=comma] {plot_data/traffic_hourly_dataset_dropout_0.csv};
            \addplot[fill=Dark2-A, fill opacity=0.3, draw=none, forget plot] fill between[of=max and min];
            \addplot[mark=None, color=Dark2-A, mark size=2pt, line width=1.5pt]
                table [x=message_loss_prob, y=mean_loss, col sep=comma] {plot_data/traffic_hourly_dataset_dropout_0.csv};

            \addplot[name path=max, draw=none, forget plot]
                table [x=message_loss_prob, y=max_loss, col sep=comma] {plot_data/traffic_hourly_dataset_dropout_1.csv};
            \addplot[name path=min, draw=none, forget plot]
                table [x=message_loss_prob, y=min_loss, col sep=comma] {plot_data/traffic_hourly_dataset_dropout_1.csv};
            \addplot[fill=Dark2-B, fill opacity=0.3, draw=none, forget plot] fill between[of=max and min];
            \addplot[mark=None, color=Dark2-B, mark size=2pt, line width=1.5pt]
                table [x=message_loss_prob, y=mean_loss, col sep=comma] {plot_data/traffic_hourly_dataset_dropout_1.csv};

            \addplot[name path=max, draw=none, forget plot]
                table [x=message_loss_prob, y=max_loss, col sep=comma] {plot_data/traffic_hourly_dataset_dropout_5.csv};
            \addplot[name path=min, draw=none, forget plot]
                table [x=message_loss_prob, y=min_loss, col sep=comma] {plot_data/traffic_hourly_dataset_dropout_5.csv};
            \addplot[fill=Dark2-C, fill opacity=0.3, draw=none, forget plot] fill between[of=max and min];
            \addplot[mark=None, color=Dark2-C, mark size=2pt, line width=1.5pt]
                table [x=message_loss_prob, y=mean_loss, col sep=comma] {plot_data/traffic_hourly_dataset_dropout_5.csv};

            \addplot[name path=max, draw=none, forget plot]
                table [x=message_loss_prob, y=max_loss, col sep=comma] {plot_data/traffic_hourly_dataset_dropout_10.csv};
            \addplot[name path=min, draw=none, forget plot]
                table [x=message_loss_prob, y=min_loss, col sep=comma] {plot_data/traffic_hourly_dataset_dropout_10.csv};
            \addplot[fill=Dark2-D, fill opacity=0.3, draw=none, forget plot] fill between[of=max and min];
            \addplot[mark=None, color=Dark2-D, mark size=2pt, line width=1.5pt]
                table [x=message_loss_prob, y=mean_loss, col sep=comma] {plot_data/traffic_hourly_dataset_dropout_10.csv};

        \nextgroupplot[
            title={London-smart-meters},
        ]
                        \addplot[name path=max, draw=none, forget plot]
                table [x=message_loss_prob, y=max_loss, col sep=comma] {plot_data/london_smart_meters_dataset_dropout_0.csv};
            \addplot[name path=min, draw=none, forget plot]
                table [x=message_loss_prob, y=min_loss, col sep=comma] {plot_data/london_smart_meters_dataset_dropout_0.csv};
            \addplot[fill=Dark2-A, fill opacity=0.3, draw=none, forget plot] fill between[of=max and min];
            \addplot[mark=None, color=Dark2-A, mark size=2pt, line width=1.5pt]
                table [x=message_loss_prob, y=mean_loss, col sep=comma] {plot_data/london_smart_meters_dataset_dropout_0.csv};

            \addplot[name path=max, draw=none, forget plot]
                table [x=message_loss_prob, y=max_loss, col sep=comma] {plot_data/london_smart_meters_dataset_dropout_1.csv};
            \addplot[name path=min, draw=none, forget plot]
                table [x=message_loss_prob, y=min_loss, col sep=comma] {plot_data/london_smart_meters_dataset_dropout_1.csv};
            \addplot[fill=Dark2-B, fill opacity=0.3, draw=none, forget plot] fill between[of=max and min];
            \addplot[mark=None, color=Dark2-B, mark size=2pt, line width=1.5pt]
                table [x=message_loss_prob, y=mean_loss, col sep=comma] {plot_data/london_smart_meters_dataset_dropout_1.csv};

            \addplot[name path=max, draw=none, forget plot]
                table [x=message_loss_prob, y=max_loss, col sep=comma] {plot_data/london_smart_meters_dataset_dropout_5.csv};
            \addplot[name path=min, draw=none, forget plot]
                table [x=message_loss_prob, y=min_loss, col sep=comma] {plot_data/london_smart_meters_dataset_dropout_5.csv};
            \addplot[fill=Dark2-C, fill opacity=0.3, draw=none, forget plot] fill between[of=max and min];
            \addplot[mark=None, color=Dark2-C, mark size=2pt, line width=1.5pt]
                table [x=message_loss_prob, y=mean_loss, col sep=comma] {plot_data/london_smart_meters_dataset_dropout_5.csv};

            \addplot[name path=max, draw=none, forget plot]
                table [x=message_loss_prob, y=max_loss, col sep=comma] {plot_data/london_smart_meters_dataset_dropout_10.csv};
            \addplot[name path=min, draw=none, forget plot]
                table [x=message_loss_prob, y=min_loss, col sep=comma] {plot_data/london_smart_meters_dataset_dropout_10.csv};
            \addplot[fill=Dark2-D, fill opacity=0.3, draw=none, forget plot] fill between[of=max and min];
            \addplot[mark=None, color=Dark2-D, mark size=2pt, line width=1.5pt]
                table [x=message_loss_prob, y=mean_loss, col sep=comma] {plot_data/london_smart_meters_dataset_dropout_10.csv};
            
        \end{groupplot}
    \end{tikzpicture}
\caption{Test Loss of Models Versus Message Loss.
    \capt{We train transformer models with \SI{0}{\percent} (\ref{ref:zero}), \SI{1}{\percent} (\ref{ref:one}), \SI{5}{\percent} (\ref{ref:five}) and \SI{10}{\percent} (\ref{ref:ten}) \ac{dropout} probability. 
    Models trained without \ac{dropout} are highly sensitive to message loss, whereas models trained with \ac{dropout} are robust, performing best near the message loss probabilities they were trained for.
    }}
    \label{fig:dropout-comparison}
\end{figure}
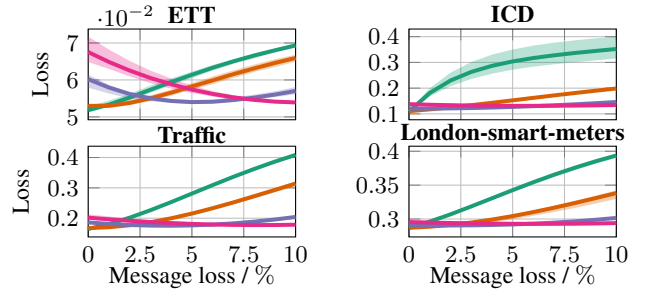

\cref{fig:dropout-comparison} analyzes \ac{dropout} against message loss. 
Without \ac{dropout}, test accuracy degrades rapidly as message loss increases, whereas \ac{dropout}-trained transformers are markedly more robust and perform best when training and deployment message-loss probabilities match.
They still fall short of the original, loss-free network: test losses increase by about \SI{4.2}{\percent} for ETT, \SI{23.6}{\percent} for ICD, \SI{8.5}{\percent} for Traffic and \SI{2.5}{\percent} for London-smart-meters, which we attribute to unrecoverable loss in the final layers.
Nonetheless, \ac{dropout} substantially improves inference-time robustness to message loss.

\section{Conclusions}
In this work, we presented \ac{name}, a framework that enables \ac{dti} on ultra-low-power wireless devices in mesh networks. 
We co-designed partitioning, communication and training mechanisms that reduce per-device \ac{ram}, flash, and compute while remaining communication-efficient.
This allows multiple devices to collaboratively execute transformer models several times larger than what any single device can sustain, as shown in our experiments. 
Our results further show that pruning inside \ac{somegather} yields substantial \ac{ram}, communication and latency savings while maintaining accuracy, and that \ac{dropout} significantly improves robustness to packet losses in wireless networks. 
However, these gains come with some trade-offs that motivate future work.

First, as each attention head is currently mapped to a single device, the number of heads upper-bounds the number of supported devices. 
Future work could distribute single heads across multiple devices, to further improve scalability.
Second, our pruning and dropout mechanisms are trained for a fixed deployment configuration (number of devices, expected message loss), which is less practical for expensive foundation models reused across diverse settings.
An important direction for future research is to equip pre-trained transformers with \ac{name}-like communication awareness without retraining for each network configuration.

Taken together, our results show that \ac{dti} is feasible ``beyond the edge'' on ultra-low-power wireless devices with tight resource budgets, paving the way for more capable 
systems built from highly constrained devices.

\clearpage

\section*{Acknowledgements}
This work was supported by the German Research Foundation (DFG) within the priority program 1914 (grant TR 1433/2) and within the Emmy Noether project NextIoT (ZI 1635/2-1), and by the LOEWE initiative (Hesse, Germany) within the emergenCITY center (LOEWE/1/12/519/03/05.001(0016)/72). 

The authors gratefully acknowledge the computing time provided to them at the NHR Center NHR4CES at RWTH Aachen University (p0021919). 

We thank Andrés Posada Moreno, Lukas Wildberger, Paul Brunzema, Paul Kruse and Fabian Mager for insightful discussions.

\bibliographystyle{named}
\bibliography{references}

\end{document}